\documentclass[letterpaper]{article} % DO NOT CHANGE THIS
\usepackage[draft]{aaai2027}  % DO NOT CHANGE THIS
\usepackage[hyphens]{url}  % DO NOT CHANGE THIS
\usepackage{graphicx} % DO NOT CHANGE THIS
\usepackage{natbib}  % DO NOT CHANGE THIS AND DO NOT ADD ANY OPTIONS TO IT
\usepackage{caption} % DO NOT CHANGE THIS AND DO NOT ADD ANY OPTIONS TO IT
\usepackage{amsmath}
\usepackage{amssymb}
\usepackage{booktabs}
\definecolor{hlgreen}{cmyk}{0.20,0.00,0.20,0.00}
\definecolor{hlred}{cmyk}{0.00,0.25,0.25,0.00}
\usepackage{soul} % for \hl and \sethlcolor
\usepackage{multirow}

\newcommand{\xmark}{\ensuremath{\times}}
\title{ALTER: Modeling Longitudinal Changes via Regional Differencing for 3D CT Report Generation}

\author{
    Dongchen Li\textsuperscript{\rm 1}
    \thanks{These authors contributed equally.},
    Jitao Liang\textsuperscript{\rm 1}\footnotemark[1],
    Wei Li\textsuperscript{\rm 2,3}
    \thanks{Corresponding author.}
}

\affiliations{
    \textsuperscript{\rm 1} College of Computer Science and Engineering,
    Northeastern University, Shenyang, China\\
    \textsuperscript{\rm 2} National Frontiers Science Center for Industrial
    Intelligence and Systems Optimization, Shenyang, China\\
    \textsuperscript{\rm 3} Key Laboratory of Intelligent Computing in Medical
    Image (MIIC), Northeastern University, Shenyang, China\\

}

\begin{document}

\maketitle

\begin{abstract}
Computed tomography (CT) is widely used for clinical diagnosis and longitudinal follow-up, yet automatically generating accurate and complete radiology reports from three-dimensional (3D) CT remains challenging. Existing methods improve fine-grained correspondence between images and text by modeling anatomical regions, but remain centered on the current examination. Consequently, patient-specific longitudinal changes within individual regions remain insufficiently modeled. Meanwhile, interval changes are often distributed across multiple anatomical regions, complicating a coherent assessment of the overall longitudinal state. We propose \textbf{A}natomically \textbf{L}ocalized \textbf{T}emporal \textbf{E}vidence \textbf{R}epresentation (ALTER) to address these limitations. Global Prior Integration (GPI) incorporates the prior CT and report to establish historical context for the current examination. Regional Proxy Differencing (RPD) enables each current anatomical region to retrieve a historical proxy from a single shared encoding of the prior volume and to derive localized interval evidence. Interval Change Fusion (ICF) further combines current abnormality states with region-distributed differences, converting their joint representation into change-aware soft prompts that guide report generation. ALTER achieves state-of-the-art results on most evaluation metrics across the RadGenome-ChestCT validation and CTRG-Chest-548K test sets. Code and data preprocessing details are available at
\url{https://github.com/peytonkarlie/ALTER/tree/main}.
\end{abstract}

% ===================== INTRODUCTION =====================
\section{Introduction}

\begin{figure}[t]
\centering
    \includegraphics[width=\columnwidth]{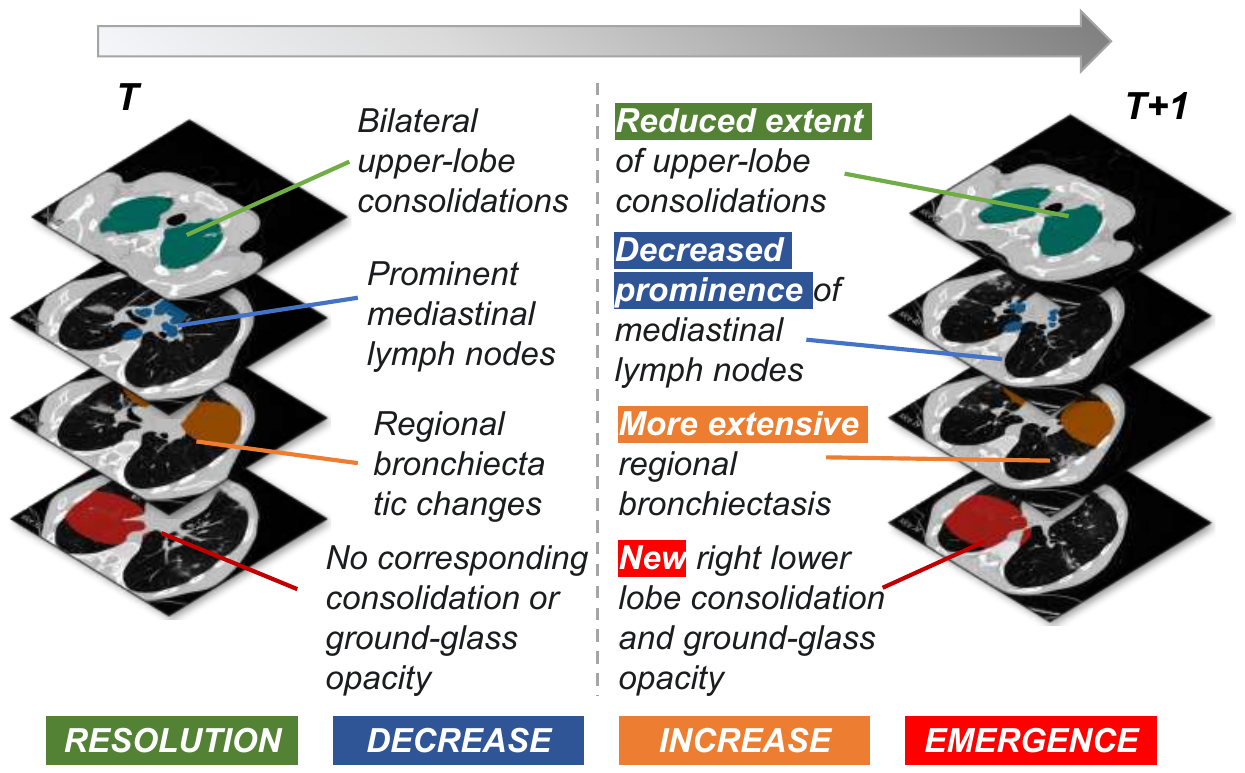}
\caption{Region-wise comparison of consecutive CT examinations reveals subtle and heterogeneous longitudinal changes across anatomical regions.}
\label{fig:motivation}
\end{figure}

Computed tomography (CT) is widely used in disease diagnosis, treatment planning, and longitudinal follow-up. Radiology reports serve as the primary medium for communicating CT findings, requiring radiologists to translate complex observations from volumetric images into coherent descriptions of anatomical structures, abnormal findings, and diagnostic impressions. In clinical practice, radiologists must not only identify abnormalities in the current examination but also compare them with patient-specific prior scans and reports to characterize interval changes and assess disease progression. This process is particularly demanding because a single CT examination typically comprises hundreds of slices and may contain multiple clinically relevant findings distributed across different anatomical regions. Radiologists must therefore review extensive volumes of images, retrieve relevant historical evidence, and synthesize current and prior findings into a temporally consistent report, resulting in a substantial clinical workload. To alleviate this burden, automatic 3D CT report generation has recently emerged as an important research direction, aiming to identify clinically meaningful findings from volumetric images and translate them into coherent radiology reports \citep{ct2rep}.

Early radiology report generation focused primarily on 2D chest radiographs, with subsequent memory-augmented Transformers and large language models improving semantic modeling and report fluency \citep{r2gen,r2gengpt}. As the task has extended to 3D CT volumes, recent methods have moved from whole-volume encoding toward anatomical regions and localized abnormalities to preserve fine-grained spatial information \citep{reg2rg, zhou2026mitigating}. However, these methods generally remain centered on the current examination and describe where findings occur without explicitly modeling how they have changed over time. Longitudinal report generation has incorporated prior images, reports, or both, but has been developed predominantly for 2D chest radiographs \citep{prefillrrg,hergen}. A small number of recent studies have begun to extend longitudinal report generation to 3D CT \citep{ct2rep,tian2026diffvpdifferentialvisualsemantic}. Even in these studies, prior examinations are incorporated through global interaction or generic discrepancy modeling, rather than by explicitly retrieving historical evidence corresponding to each current anatomical region. Consequently, subtle region-specific changes may be diluted, or distinct anatomical trajectories may be conflated.

This limitation becomes more pronounced when multiple findings evolve differently across a CT volume. Their changes may vary in direction, magnitude, and clinical relevance, making it difficult to derive a coherent longitudinal description from coarse historical context alone. Moreover, findings in the current examination do not by themselves reveal whether an abnormality is new, persistent, resolving, or worsening, while a single global comparison may blur distinct trajectories across anatomical locations. Figure~\ref{fig:motivation} illustrates that findings across anatomical regions can follow distinct trajectories between consecutive CT examinations, whereas coarse whole-examination comparison may fail to distinguish such region-specific changes. Consequently, existing models may overlook subtle interval changes or reduce heterogeneous disease evolution to an overly coarse longitudinal progression.

To address these limitations, we propose Anatomically Localized Temporal Evidence Representation (ALTER), a framework for longitudinal 3D CT report generation. Given a current CT volume, its anatomical regions, and, when available, a prior CT--report pair, ALTER coordinates three complementary mechanisms: prior integration, regional differencing, and interval change fusion. The central design of ALTER proceeds from establishing study-level historical context, through deriving region-specific interval evidence, to fusing distributed changes with current abnormality states for report generation. For study-wide historical conditioning, Global Prior Integration (GPI) first interacts current and prior volume representations through gated attention and then incorporates the encoded prior report, allowing complementary visual and textual evidence to contextualize the current examination. For localized temporal comparison, Regional Proxy Differencing (RPD) uses each current anatomical region to query a historical proxy from a single shared encoding of the prior volume. It subsequently constructs both an implicit difference enhancement and an explicit regional difference token, preserving localized change evidence without separately encoding or registering prior regional crops. Finally, Interval Change Fusion (ICF) combines the current abnormality profile with difference evidence distributed across anatomical regions and maps the resulting representation into change-aware soft prompts for the report decoder. A history availability mask disables operations that require prior examinations, enabling the same framework to process studies both with and without accessible patient history.

Our contributions are threefold:
\begin{itemize}
    \item We introduce ALTER, a framework that extends anatomy-aware 3D CT reporting to patient-specific longitudinal modeling by coordinating multimodal historical context, anatomy-specific temporal comparison, and report-level change aggregation, while supporting studies both with and without accessible patient history.

    \item We develop Regional Proxy Differencing to derive localized interval representations from a single shared encoding of the prior volume, together with Interval Change Fusion to consolidate distributed regional differences and current abnormality states into change-aware prompts for report generation.

    \item We conduct comprehensive experiments on two public 3D CT benchmarks, including component ablations, prior availability analyses, and conditioning diagnostics, to evaluate the effectiveness and complementarity of the proposed design.
\end{itemize}

% ===================== RELATED WORK =====================
\section{Related Work}
\subsection{Vision--Language Alignment}
Vision--language alignment establishes semantic correspondence between visual representations and natural-language descriptions, requiring radiology models to map subtle imaging findings to clinically precise expressions while preserving anatomical context. Early encoder--decoder approaches use co-attention or joint image--text representations to connect radiographic patterns with report semantics \citep{jing2018,tienet}. Cross-modal memory networks and curriculum learning subsequently improve the retention and acquisition of aligned clinical information \citep{r2gencmn,cmcl}. Knowledge-guided methods incorporate retrieved templates, structured graphs, and prior clinical knowledge to organize report content, while posterior and prior knowledge distillation transfers clinically relevant semantics from auxiliary sources \citep{kerp,kgrrg,ppked}. More recent studies establish fine-grained correspondence among localized visual features, disease concepts, and anatomical structures through hierarchical or region-aware modeling \citep{aligntransformer,rgrg,reg2rg}. Detected regions and region-referenced decoding further provide explicit spatial constraints for grounding individual findings in their corresponding anatomical structures. For volumetric CT, pathological graphs model spatially distributed abnormalities and their clinical relationships \citep{pathgraphbrainct}. Despite these advances, existing methods mainly address spatial image--text alignment within individual examinations, while longitudinal correspondence between the same anatomical regions across successive examinations remains underexplored.

\subsection{Longitudinal Report Generation}
Longitudinal report generation incorporates patient-specific history, including prior images, reports, or both, to describe interval changes and generate the current report. Comparison-prior and pre-filling methods leverage historical reports or draft content to provide contextual findings and guide subsequent generation \citep{comparisonprior,prefillrrg,hcllm}. Recent work further strengthens the utilization of patient-specific history through contrastive pre-training and coarse-to-fine decoding \citep{priorrg}, while multimodal analogical reasoning explicitly characterizes disease evolution across examinations \citep{mare}. Temporal representation learning methods model relationships between paired chest radiographs through contrastive objectives, longitudinal encoders, or cross-examination feature interactions \citep{bannur2023learning,controllong,hergen}. Temporal captioning and longitudinal semantic rewards further encourage the learned representations and generated reports to remain consistent with paired examinations \citep{cocacxr,ssreward}. Other approaches explicitly model disease progression or visual residuals to emphasize changes between consecutive examinations \citep{recap,ddatr}. For volumetric imaging, CT2Rep extends longitudinal modeling to 3D CT report generation by incorporating prior-image cross-attention and memory mechanisms \citep{ct2rep}. DiffVP similarly encodes global and local scan-to-reference discrepancies as visual prefixes for language-model-based CT report generation \citep{tian2026diffvpdifferentialvisualsemantic}. Nevertheless, most longitudinal systems are designed for two-dimensional radiographs, whereas existing studies on 3D CT remain limited and generally represent temporal context only at the examination level. Such global comparison may dilute subtle changes confined to individual anatomical structures or conflate the distinct trajectories of different regions. Consequently, fine-grained longitudinal comparison of the same anatomical regions across volumetric CT examinations remains insufficiently explored.

% ===================== METHOD =====================
\section{ALTER}

ALTER generates an anatomically grounded report for the current examination by jointly interpreting the current CT, its anatomical regions, and the preceding examination. Let $X_t$ and $Y_t=(y_{t,1},\ldots,y_{t,|Y_t|})$ denote the current volume and report, respectively, and let $\{X_t^r\}_{r=1}^{R}$ denote the $R$ anatomical crops extracted from the current CT. The paired history comprises the prior volume $X_{t-1}$ and report $Y_{t-1}$. Given $\mathcal{I}_t=(X_t,\{X_t^r\}_{r=1}^{R},X_{t-1},Y_{t-1})$, report generation is formulated as
\begin{equation}
 p_\theta(Y_t\mid\mathcal{I}_t)
 =\prod_{j=1}^{|Y_t|}p_\theta(y_{t,j}\mid y_{t,<j},\mathcal{I}_t).
\label{eq:harp_task}
\end{equation}

\begin{figure*}[t]
\centering
    \includegraphics[width=\linewidth]{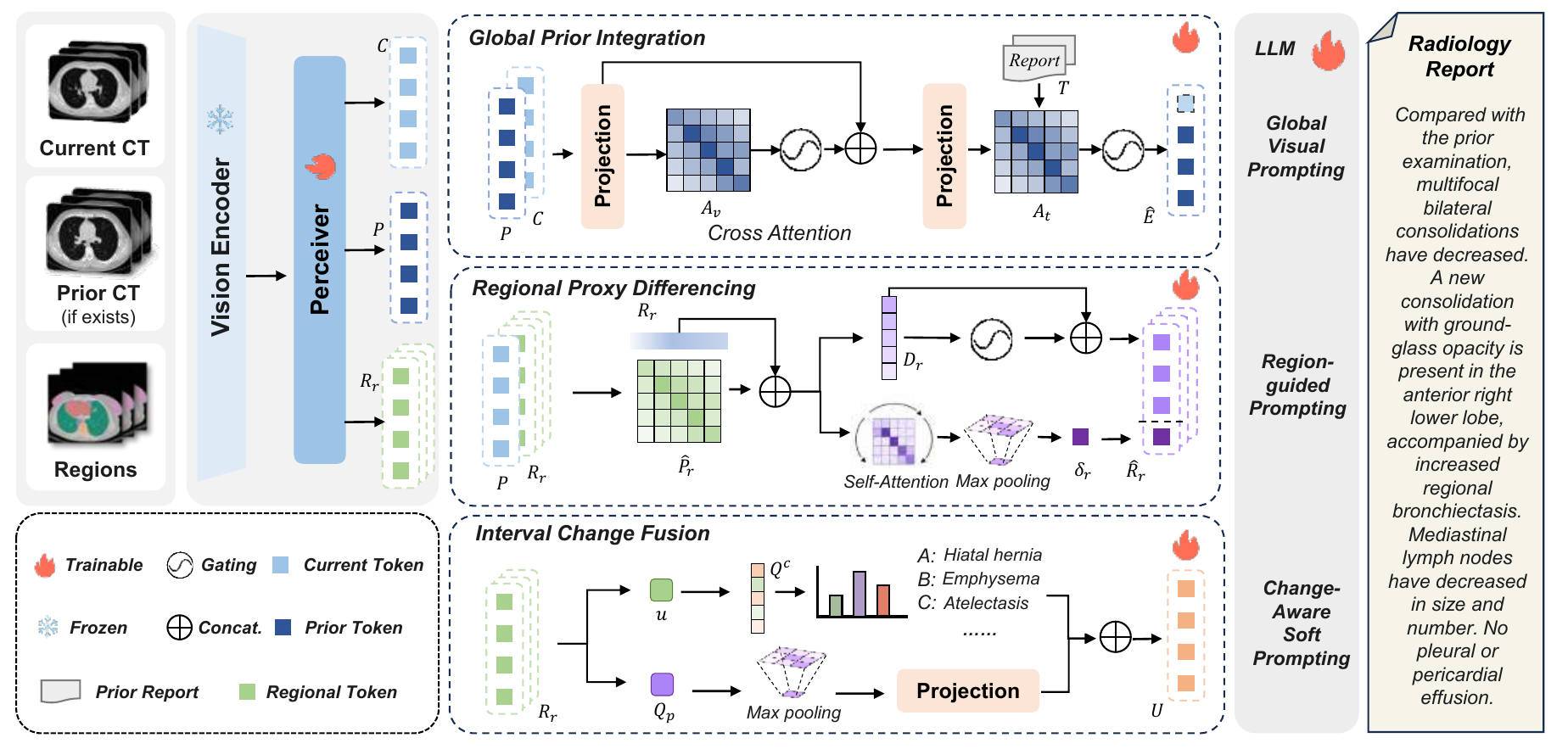}
\caption{Overview of ALTER. GPI integrates study-level historical context, RPD derives region-specific interval evidence, and ICF converts regional changes and current abnormality states into change-aware prompts for report generation.}
\label{fig:overview}
\end{figure*}

A shared Vision Transformer (ViT) and Perceiver resampler encode the current and prior volumes into study-level tokens $\mathbf{C},\mathbf{P}\!\in\!\mathbb{R}^{M\times d_v}$, respectively, and each current crop into regional tokens $\mathbf{R}_r\!\in\!\mathbb{R}^{N\times d_v}$. Here, $M$ and $N$ denote the numbers of study-level and regional tokens, respectively, while $d_v$ and $d_\ell$ denote the visual and decoder dimensions. As illustrated in Figure~\ref{fig:overview}, ALTER organizes longitudinal evidence at three complementary levels. GPI establishes study-level historical context from the prior CT and report, RPD resolves prior visual information into anatomy-specific proxies and localized longitudinal differences, and ICF consolidates the current abnormality profile with changes distributed across anatomical regions. Consequently, the decoder receives a global historical representation, region-referenced visual sequences containing dedicated difference tokens, and a compact change-aware prompt. This organization preserves the anatomical structure of CT reporting while assigning distinct and complementary roles to each historical source during generation.

Throughout this section, $\mathcal{A}(\mathbf{Q},\mathbf{S})$ denotes a pre-normalized multi-head cross-attention block in which $\mathbf{Q}$ provides the queries and $\mathbf{S}$ provides both the keys and values. Standard attention projections, residual connections, normalization operations, and padding masks are omitted below to emphasize the proposed data flow. Symbols $\Phi_{\cdot}$ and $F_{\cdot}$ denote learned affine projections and two-layer networks with sigmoid linear unit (SiLU) activations, respectively.

\subsection{Global Prior Integration}
\label{sec:gpi}

GPI constructs a fixed-length study-level representation through two sequential interactions with historical information. In each interaction, the current representation serves as the query that determines the requested information, while the corresponding historical source provides complementary context. Given a query sequence $\mathbf{Q}$ and a historical source sequence $\mathbf{S}$, we define a gated residual interaction as
\begin{equation}
\mathcal{H}_{\psi}(\mathbf{Q},\mathbf{S})
=\mathbf{Q}+\mathbf{g}_{\psi}(\mathbf{Q},\mathbf{S})
\odot\mathcal{A}(\mathbf{Q},\mathbf{S}),
\label{eq:gpi_history_operator}
\end{equation}
where $\mathbf{g}_{\psi}(\mathbf{Q},\mathbf{S})=\sigma(F_{\psi}([\mathbf{Q}\Vert\mathcal{A}(\mathbf{Q},\mathbf{S})]))$ denotes a feature-wise gate, and $\Vert$ and $\odot$ denote concatenation and element-wise multiplication, respectively. The residual path maintains the current examination as the representational reference, while the gate restricts the influence of historical information to feature dimensions supported by the retrieved evidence.

\paragraph{Visual history.}
Learned temporal embeddings $\mathbf{e}_t$ and $\mathbf{e}_{t-1}$ distinguish the current and prior examinations, yielding $\mathbf{C}^{t}=\mathbf{C}+\mathbf{e}_t$ and $\mathbf{P}^{t-1}=\mathbf{P}+\mathbf{e}_{t-1}$. The current tokens query the prior sequence to obtain a historical response aligned with the current token positions, $\mathbf{A}_v=\mathcal{A}(\mathbf{C}^{t},\mathbf{P}^{t-1})$. The resulting gated fusion is
\begin{equation}
\mathbf{G}=\mathcal{H}_{v}(\mathbf{C}^{t},\mathbf{P}^{t-1}).
\label{eq:gpi_visual_fusion}
\end{equation}
Here, $\mathbf{A}_v$ aligns relevant information from the prior volume with the current token positions, while $\mathbf{G}$ selectively incorporates this information through the gate. Because the interaction operates on compact study-level sequences, it captures examination-level longitudinal correspondence without requiring explicit voxel-level registration.

\paragraph{Report history.}
The visually contextualized sequence is first projected into the decoder space as $\mathbf{E}=\Phi_c(\mathbf{G})\!\in\!\mathbb{R}^{M\times d_\ell}$. In parallel, a frozen text encoder and a learned projection map the prior report to $\mathbf{T}\!\in\!\mathbb{R}^{L\times d_\ell}$. The aligned textual response is then computed as $\mathbf{A}_t=\mathcal{A}(\mathbf{E},\mathbf{T})$. Applying the same asymmetric gated interaction yields
\begin{equation}
\hat{\mathbf{E}}=\mathcal{H}_{t}(\mathbf{E},\mathbf{T}).
\label{eq:gpi_text_fusion}
\end{equation}
A report mask excludes padded positions from the interaction. The resulting $\hat{\mathbf{E}}$ therefore serves as a study-level prompt in which the prior report is incorporated only after the current examination has been contextualized by the prior volume, thereby maintaining a clear temporal reference across modalities. This ordering allows the textual history to refine a visual representation already anchored to both examinations without directly governing the visual comparison.

\subsection{Regional Proxy Differencing}
\label{sec:rpd}

RPD recovers anatomy-specific historical information from the shared prior-volume sequence $\mathbf{P}$. Whole-volume differencing would expose every anatomical region to the same historical summary, whereas longitudinal findings require a distinct comparison for each current region. RPD therefore uses the current regional representation to identify the relevant information within the shared prior sequence before constructing the corresponding difference representation. The retrieved counterpart is a soft proxy rather than an explicitly registered crop, enabling localized comparison without repeated encoding of the prior volume or explicit regional registration.

\paragraph{Regional prior retrieval.}
For each region $r$, the current regional tokens $\mathbf{R}_r$ query the prior study tokens, and an output projection preserves the regional token layout:
\begin{equation}
\tilde{\mathbf{P}}_r=\Phi_o\!\left(\mathcal{A}(\mathbf{R}_r,\mathbf{P})\right),
\label{eq:rpd_proxy}
\end{equation}
where $\Phi_o$ is shared across regions. The resulting proxy $\tilde{\mathbf{P}}_r\!\in\!\mathbb{R}^{N\times d_v}$ has the same token layout as $\mathbf{R}_r$, while its content is retrieved from $\mathbf{P}$ through an anatomy-specific query. Thus, all regions reuse a single encoding of the prior volume while obtaining distinct historical views.

\paragraph{Implicit regional differencing.}
The signed token difference $\mathbf{D}_r=\mathbf{R}_r-\tilde{\mathbf{P}}_r$ characterizes deviations of the current region from its retrieved historical proxy. Directly replacing the current representation with this difference, however, would discard persistent anatomical context. RPD therefore computes a gate as $\mathbf{z}_r=\sigma(F_z([\mathbf{R}_r\Vert\tilde{\mathbf{P}}_r\Vert\mathbf{D}_r]))$ and decomposes the representation update into a gated change residual and a moderated proxy residual:
\begin{equation}
\hat{\mathbf{R}}_r=
\mathbf{R}_r+\mathbf{z}_r\odot\Phi_\Delta(\mathbf{D}_r)
+\alpha\Phi_p(\tilde{\mathbf{P}}_r),
\label{eq:rpd_fusion}
\end{equation}
where $\alpha=1/2$ controls the direct contribution of the historical proxy. The gated change residual selectively emphasizes features associated with longitudinal deviations, the proxy residual retains context shared across examinations, and the outer current-token residual keeps the updated representation anchored to the current anatomy being reported.

\paragraph{Explicit regional difference.}
The implicit change information remains distributed across $N$ regional tokens and may therefore be difficult for the decoder to access as a concise comparison cue. RPD further refines the pre-fusion difference as $\mathbf{H}_r=\mathcal{A}(\mathbf{D}_r,\mathbf{D}_r)$ and then extracts a dedicated regional difference token:
\begin{equation}
\boldsymbol{\delta}_r=\Phi_\delta\!\left(\operatorname{MaxPool}_N(\mathbf{H}_r)\right).
\label{eq:rpd_explicit}
\end{equation}
Feature-wise max pooling along the regional token dimension preserves the strongest difference response in each feature dimension, producing $\boldsymbol{\delta}_r\!\in\!\mathbb{R}^{1\times d_v}$. After being projected to $d_\ell$, this token is appended to the corresponding regional sequence together with its anatomical mask embedding. The decoder thus receives two complementary representations of the same regional comparison: $\hat{\mathbf{R}}_r$ preserves detailed current anatomy with distributed historical modulation, whereas $\boldsymbol{\delta}_r$ explicitly provides a compact summary of region-specific changes.

\subsection{Interval Change Fusion}
\label{sec:ICF}

ICF constructs $n$ report-level soft prompt tokens from two complementary signals: the current abnormality profile and visual changes distributed across anatomical regions. The status pathway identifies which abnormalities should be described, while the change pathway summarizes how their regional manifestations differ from the prior examination. Mapping both pathways to the same prompt positions provides the decoder with a compact examination-level representation while preserving the regional features used for localized generation.

\paragraph{Abnormality status.}
The status pathway first aggregates the difference-enhanced regional sequences:
\begin{equation}
\mathbf{u}=\Phi_e\!\left(
\frac{1}{R}\sum_{r=1}^{R}\operatorname{mean}_N(\hat{\mathbf{R}}_r)
\right).
\label{eq:ICF_pool}
\end{equation}
We use $\operatorname{sg}(\cdot)$ to denote gradient stopping. A lightweight classifier maps $\operatorname{sg}(\mathbf{u})$ to logits $\boldsymbol{\ell}\!\in\!\mathbb{R}^{K\times2}$ for $K=18$ abnormalities, and their positive-class probabilities form $\mathbf{q}\!\in\![0,1]^K$. During teacher-forced training, the conditioning vector $\mathbf{q}^{c}$ uses report-derived pseudo labels $\mathbf{y}^{\ast}$; at inference, it uses the predicted probabilities $\mathbf{q}$. Let $\mathcal{P}_s:\mathbb{R}^{K}\!\rightarrow\!\mathbb{R}^{n\times d_\ell}$ denote a normalized multilayer projection into the decoder prompt space. The status prompt is defined as
\begin{equation}
\mathbf{S}=\mathbf{B}_s+\gamma_s\mathcal{P}_s(\mathbf{q}^{c}),
\label{eq:ICF_status_prompt}
\end{equation}
where $\mathbf{B}_s\!\in\!\mathbb{R}^{n\times d_\ell}$ is a learned base prompt and $\gamma_s$ is a learned scalar. The base prompt establishes the soft-token positions, while the scaled residual introduces examination-specific abnormality semantics.

\paragraph{Distributed change evidence.}
The regional difference tokens are stacked as $\boldsymbol{\Delta}=\operatorname{stack}(\boldsymbol{\delta}_1,\ldots,\boldsymbol{\delta}_R)\!\in\!\mathbb{R}^{R\times d_v}$. Learnable queries $\mathbf{Q}_c\!\in\!\mathbb{R}^{n\times d_v}$ retrieve complementary regional change patterns as $\mathbf{A}_c=\mathcal{A}(\mathbf{Q}_c,\operatorname{sg}(\boldsymbol{\Delta}))$, and their mean response is denoted by $\bar{\mathbf{a}}_c=\operatorname{mean}_n(\mathbf{A}_c)$. Let $\mathcal{P}_c:\mathbb{R}^{d_v}\!\rightarrow\!\mathbb{R}^{n\times d_\ell}$ denote the corresponding normalized projection into the prompt space. The change prompt is defined as
\begin{equation}
\mathbf{Z}=\gamma_c\mathcal{P}_c(\bar{\mathbf{a}}_c).
\label{eq:ICF_change}
\end{equation}
The scalar $\gamma_c$ controls the contribution of the distributed change evidence. Averaging the query responses produces a shared examination-level change summary, which $\mathcal{P}_c$ maps to the $n$ prompt positions. The final prompt $\mathbf{U}=\mathbf{S}+\mathbf{Z}$ aligns current abnormality semantics with region-distributed change cues before being prepended to the decoder input. Distinct queries retrieve complementary combinations of regional changes, while mapping their aggregate to the same $n$ positions as $\mathbf{S}$ makes the status and change evidence directly composable without increasing the decoder context length.

\paragraph{Training objective.}
The report-derived pseudo labels both supervise the abnormality classifier and provide status conditioning during training. For a batch of $B$ studies, the language modeling (LM) objective and auxiliary status objective are jointly optimized:
\begin{equation}
\mathcal{L}
=\mathcal{L}_{\mathrm{LM}}+\frac{\lambda_s}{BK}
\sum_{i=1}^{B}\sum_{k=1}^{K}
\operatorname{CE}(\boldsymbol{\ell}_{ik},y_{ik}^{\ast}).
\label{eq:ICF_objective}
\end{equation}
Here, $\operatorname{CE}$ denotes cross-entropy, and $\lambda_s=0.1$. Gradient stopping confines direct auxiliary supervision to the status head, preventing either prompt pathway from modifying the regional visual features through this auxiliary objective. Soft-prompt positions are excluded from the token-level LM targets, while $\mathcal{L}_{\mathrm{LM}}$ optimizes the prompt projections and low-rank adaptation (LoRA) parameters through report generation \citep{lora}.

% ===================== EXPERIMENTS =====================
\section{Experiments}

\subsection{Datasets}
We evaluate ALTER on two complementary chest CT benchmarks. \textbf{RadGenome-ChestCT} \citep{radgenome}, derived from CT-RATE \citep{hamamci2026generalist}, provides anatomical region masks and region-grounded descriptions and serves as our primary benchmark for region-aware report generation and evaluation stratified by prior availability. \textbf{CTRG-Chest-548K} \citep{ctrg} is an independent Chinese-language corpus used to evaluate robustness across datasets and reporting styles. To adapt it to the same regional reporting interface, we employ an AI-assisted pipeline to process the original reports into anatomy-specific descriptions. Detailed preprocessing procedures, split statistics, prior-selection strategies, and CTRG report processing are provided in the \emph{Dataset Details} section of the appendix.

\subsection{Evaluation Metrics}
We report three complementary families of metrics to evaluate textual quality and clinical correctness. First, NLG metrics include BLEU-1--4 (B-1--B-4) \citep{bleu}, ROUGE-L (R-L) \citep{rouge}, METEOR (M) \citep{meteor}, and CIDEr \citep{cider}. These metrics assess textual similarity from complementary perspectives, including n-gram precision, sequence overlap, flexible matching, and report-specific expressions. Second, clinical efficacy metrics include CE-P, CE-R, and CE-F1. Specifically, a fixed labeler extracts 18-class abnormality labels from the generated and reference reports, and the metrics are computed by comparing the two label sets \citep{hamamci2026generalist}. These metrics quantify the extent to which clinically relevant abnormalities are preserved in the generated reports. Third, clinical-semantic metrics include RadGraph F1 (RG) and GREEN. RadGraph F1 measures structured semantic consistency by comparing clinical entities and their relations between generated and reference reports \citep{radgraph}, whereas GREEN employs a radiology-oriented LLM evaluator to assess overall clinical correctness by identifying important omissions, unsupported findings, and errors in location or severity \citep{green}.

\subsection{Baselines}
We compare ALTER with seven representative baselines spanning complementary research directions in medical report generation. R2GenGPT \citep{r2gengpt} represents LLM-based radiology report generation, whereas MedVInT \citep{medvint} represents medical visual instruction tuning. M3D \citep{bai2024m3dadvancing3dmedical} and RadFM \citep{radfm} are generalist medical vision--language models capable of processing volumetric medical images. For specialized 3D CT report generation, we include CT2Rep \citep{ct2rep}, along with the fine-grained grounding methods DEAR \citep{zhou2026mitigating} and Reg2RG \citep{reg2rg}, which model entity-level and anatomical region-level correspondence, respectively. Reg2RG is the most closely related baseline because it likewise employs anatomy-aware regional representations, whereas ALTER further extends region-aware modeling to patient-specific longitudinal comparison. Comparisons with these baselines allow us to assess whether patient-specific longitudinal modeling provides additional benefits over existing volumetric and anatomy-aware report generation approaches.

\subsection{Implementation Details}
We aggregate the masks generated by TotalSegmentator \citep{wasserthal2023totalsegmentator} into $R{=}10$ anatomical reporting regions. A frozen ViT \citep{dosovitskiy2020image} initialized from the RadFM 3D visual checkpoint \citep{radfm} separately encodes the full CT volume and regional inputs, while a shared Perceiver resampler \citep{alayrac2022flamingo} compresses each feature sequence into 32 latent tokens. These visual tokens are then projected into the 4096-dimensional embedding space of RadLLaMA-7B \citep{radllama7b} for report generation. Prior reports are independently encoded by frozen RadBERT-RoBERTa-4m \citep{yan2022radbert} and projected into the same embedding space. ICF employs four soft prompt tokens. We first jointly train GPI and RPD for 10 epochs using AdamW \citep{loshchilov2017decoupled} with a peak learning rate of $5\times10^{-5}$ and 100 warmup steps, followed by ICF fine-tuning. Initialized to $\gamma_s{=}0.02$ and $\gamma_c{=}0.01$, the ICF gates converge to $\gamma_s{=}0.0211$ and $\gamma_c{=}0.0109$; they remain close to their initial values, consistent with magnitude-limited ICF modulation. Numerically sensitive operations are performed in FP32, while the remaining computations use BF16 mixed precision. Further details on optimization, parameter initialization, and supervision are provided in the \emph{Model Configuration} subsection of the appendix, with efficiency analysis presented in \emph{Computational Cost}.

\subsection{Experimental Results}

\subsubsection{Main Results}
\label{sec:main_results}

\begin{table*}[t]
\centering
\small
\setlength{\tabcolsep}{3.2pt}
\begin{tabular}{l|ccc@{\;}ccc@{\;}cc|ccccc}
\toprule
 & \multicolumn{8}{c|}{\textbf{RadGenome-ChestCT}} & \multicolumn{5}{c}{\textbf{CTRG-Chest-548K}} \\
\cmidrule(lr){2-9}\cmidrule(l){10-14}
\textbf{Method} & B-4 & R-L & M & CE-P & CE-R & CE-F1 & RG & GREEN & B-4 & R-L & M & RG & GREEN \\
\midrule
R2GenGPT (2023, Meta-Rad.) & 24.2 & 32.3 & 39.9 & 34.0 & 6.6 & 11.0 & 22.2 & 33.5 & 30.1 & \underline{50.9} & 47.1 & -- & -- \\
MedVInT (2024, Commun. Med.) & 24.6 & 32.6 & 40.4 & 37.7 & 14.8 & 21.2 & -- & -- & 30.7 & 49.5 & 49.3 & -- & -- \\
RadFM (2025, Nat. Commun.) & 23.7 & 31.5 & 39.9 & 38.2 & 13.1 & 19.5 & -- & 23.4 & 30.9 & 49.1 & 49.2 & -- & -- \\
CT2Rep (2024, MICCAI) & 23.6 & 31.0 & 40.2 & 31.7 & 8.9 & 13.9 & \underline{23.2} & \textbf{37.4} & 29.2 & 50.2 & 47.0 & -- & -- \\
M3D (2024, arXiv) & 24.5 & 32.6 & 40.0 & 40.7 & 9.0 & 14.8 & 16.7 & 29.7 & 30.9 & 50.2 & 49.3 & -- & -- \\
Reg2RG (2025, TMI) & \underline{25.2} & \underline{37.2} & \underline{42.7} & \underline{49.4} & \underline{16.1} & \underline{24.3} & 20.8 & 33.7 & \underline{34.2} & 49.8 & \underline{57.8} & \underline{30.6} & \textbf{35.1} \\
DEAR (2026, AAAI) & 14.7 & 27.1 & 39.5 & 30.7 & \textbf{46.0} & \textbf{36.8} & 10.1 & 12.3 & 16.1 & 30.0 & 40.4 & 28.7 & 23.1 \\
\midrule
\textbf{ALTER} & \textbf{27.9} & \textbf{41.6} & \textbf{44.3} & \textbf{49.6} & 12.8 & 20.4 & \textbf{25.5} & \underline{36.3} & \textbf{39.1} & \textbf{53.9} & \textbf{57.9} & \textbf{32.2} & \underline{33.0} \\
\bottomrule
\end{tabular}
\normalsize
\caption{Comparison with representative CT report generation baselines on the RadGenome-ChestCT validation set and the CTRG-Chest-548K test set. Bold and underlined values indicate the best and second-best results, respectively. ``--'' indicates that the corresponding result is unavailable.}
\label{tab:main_results}
\end{table*}

Table~\ref{tab:main_results} compares ALTER with representative methods on two CT report generation benchmarks. On RadGenome-ChestCT, ALTER achieves the best results on five of the eight metrics, including all NLG metrics, CE-P, and RG. Compared with the strongest baseline for each metric, ALTER improves B-4 from 25.2 to 27.9, R-L from 37.2 to 41.6, and RG from 23.2 to 25.5. These gains cover both lexical similarity and the consistency of clinical entities and relations, showing that the benefits of ALTER extend beyond surface-level text overlap. ALTER also outperforms Reg2RG on six of the eight metrics, increasing RG from 20.8 to 25.5 and GREEN from 33.7 to 36.3. The consistent improvements over Reg2RG in RG and GREEN are meaningful because these two metrics evaluate complementary aspects of clinical fidelity. RG measures the structural consistency of clinical entities and relations, while GREEN penalizes clinically consequential omissions and unsupported statements. Consistent gains across these complementary metrics provide stronger evidence of clinically coherent and reliable generation than any single score considered in isolation. This overall advantage suggests that patient-specific longitudinal information effectively complements anatomical regional representations. For abnormality recognition, ALTER achieves the highest CE-P, whereas DEAR obtains higher CE-R and CE-F1 scores. The \emph{Abnormality Results} subsection of the appendix further analyzes this difference in the precision--recall trade-off.

On CTRG-Chest-548K, ALTER ranks first on four of the five metrics. Compared with the strongest baseline for each metric, ALTER improves B-4, R-L, and M by 4.9, 3.0, and 0.1 points, respectively, and achieves the highest RG score of 32.2. Its GREEN score of 33.0 remains below that of Reg2RG. These results indicate that the main advantages of ALTER on this dataset lie in textual consistency and the preservation of structured clinical content. Because no matched prior examination is available under our pairing protocol, this benchmark evaluates report generation without historical information rather than longitudinal reasoning. Nevertheless, the strong performance of ALTER demonstrates that the method remains effective when prior examinations are unavailable. Region-wise results are provided in the \emph{Regional Results} subsection of the appendix.

\subsubsection{Ablation Study}
\begin{table*}[t]
\centering
\small
\setlength{\tabcolsep}{3.2pt}
\begin{tabular}{lccc|cccc@{\;}ccc@{\;}cc|cccc}
\toprule
 & & & & \multicolumn{9}{c|}{\textbf{RadGenome-ChestCT}} & \multicolumn{4}{c}{\textbf{CTRG-Chest-548K}} \\
\cmidrule(lr){5-13}\cmidrule(l){14-17}
\textbf{Variant} & \textbf{GPI} & \textbf{RPD} & \textbf{ICF} & B-4 & R-L & M & CIDEr & CE-P & CE-R & CE-F1 & RG & GREEN & B-4 & R-L & M & GREEN \\
\midrule
(a) & \xmark & \xmark & \xmark & 25.2 & 37.2 & 42.7 & 7.3 & 49.4 & \textbf{16.1} & \textbf{24.3} & 20.8 & 33.7 & 34.2 & 49.8 & 57.8 & 35.1 \\
(b)  & \checkmark & \xmark & \xmark & 27.8 & 38.3 & 43.8 & 11.7 & 45.3 & 9.4 & 15.5 & 25.1 & \textbf{38.9} & 31.8 & 46.8 & 52.2 & \textbf{44.3} \\
(c)  & \checkmark & \checkmark & \xmark & \textbf{28.1} & 39.9 & \textbf{44.4} & \textbf{12.9} & 42.0 & 11.9 & 18.6 & \textbf{25.5} & 35.8 & 33.0 & 48.5 & 54.4 & 36.2 \\
(d)  & \checkmark & \checkmark & \checkmark & 27.9 & \textbf{41.6} & 44.3 & 12.3 & \textbf{49.6} & 12.8 & 20.4 & \textbf{25.5} & 36.3 & \textbf{39.1} & \textbf{53.9} & \textbf{57.9} & 33.0 \\
\bottomrule
\end{tabular}
\normalsize
\caption{Cumulative ablations for GPI, RPD, and ICF. Checkmarks and crosses indicate enabled and disabled modules, respectively.}
\label{tab:ablation_cumulative}
\end{table*}

Table~\ref{tab:ablation_cumulative} adopts a cumulative ablation design in which GPI, RPD, and ICF are introduced successively, with each configuration trained independently. On RadGenome-ChestCT, GPI yields the largest initial gains, increasing CIDEr from 7.3 to 11.7, RG from 20.8 to 25.1, and GREEN from 33.7 to 38.9. This step accounts for most of the improvement in report quality and clinical agreement, although CE-F1 decreases from 24.3 to 15.5. Adding RPD further improves B-4, R-L, CE-R, and CE-F1, and reaches the jointly highest RG score of 25.5, while GREEN drops from 38.9 to 35.8. These results indicate that RPD primarily improves abnormality coverage and the structural consistency of clinical entities and relations. Incorporating ICF in the final step achieves the best or jointly best results on three of the nine metrics, reaching 41.6 in R-L, 49.6 in CE-P, and 25.5 in RG.

On CTRG-Chest-548K, the complete configuration achieves the best or jointly best results on three of the four reported metrics. Incorporating ICF improves B-4 from 33.0 to 39.1, R-L from 48.5 to 53.9, and M from 54.4 to 57.9, whereas GREEN decreases from 36.2 to 33.0. Across both datasets, the results show that GPI, RPD, and ICF contribute to different metrics rather than yielding uniformly cumulative gains. The \emph{RPD Proxy} subsection of the appendix further compares the region-conditioned RPD proxy with global averaging of the prior representation.

\subsubsection{Longitudinal Subset Analysis}
\label{sec:longitudinal_subset}

\begin{table}[t]
\centering
\small
\setlength{\tabcolsep}{4.0pt}
\begin{tabular}{@{}clcccc@{}}
\toprule
\textbf{Set} & \textbf{Model} & \textbf{B-4} & \textbf{M} &
\textbf{R-L} & \textbf{CIDEr} \\
\midrule
\multirow[c]{2}{*}{All}
& Baseline & 25.2 & 42.7 & 37.2 & 7.3 \\
& Ours & \textbf{27.9} & \textbf{44.3} & \textbf{41.6} & \textbf{12.3} \\
\addlinespace[2pt]
\multirow[c]{2}{*}{w/ prior}
& Baseline & \textbf{22.3} & \textbf{39.0} & 33.8 & 9.5 \\
& Ours & 21.6 & 38.6 & \textbf{35.4} & \textbf{10.5} \\
\addlinespace[2pt]
\multirow[c]{2}{*}{w/o prior}
& Baseline & 25.8 & 43.4 & 37.8 & 7.1 \\
& Ours & \textbf{29.2} & \textbf{45.4} & \textbf{42.8} & \textbf{12.7} \\
\bottomrule
\end{tabular}
\normalsize
\caption{NLG results on RadGenome-ChestCT validation under three evaluation settings: the complete validation set (All), cases with an eligible prior CT (w/ prior), and cases without an eligible prior CT (w/o prior).}
\label{tab:prior_regimes}
\end{table}

Table~\ref{tab:prior_regimes} stratifies results by prior availability. For the 260 studies with priors, ALTER raises R-L from 33.8 to 35.4 and CIDEr from 9.5 to 10.5, while B-4 drops from 22.3 to 21.6 and M drops from 39.0 to 38.6. Thus, on studies with priors, ALTER improves report-level and study-specific correspondence despite slightly lower B-4 and M scores.

For the 1{,}304 studies without priors, ALTER increases B-4 from 25.8 to 29.2, M from 43.4 to 45.4, R-L from 37.8 to 42.8, and CIDEr from 7.1 to 12.7. Because history-dependent operations are disabled, these gains arise from components operating on the current examination. The subsets are reported separately because prior availability changes both model input and case composition; they should not be treated as matched cohorts. Overall, these results show that the aggregate gains come mainly from studies without priors. Paired examples are provided in the \emph{Longitudinal Cases} section of the appendix.

\subsubsection{ICF Conditioning Diagnostic}
\label{sec:ICF_conditioning}

\begin{table}[t]
\centering
\small
\setlength{\tabcolsep}{2.2pt}
\begin{tabular}{@{}lcccccc@{}}
\toprule
\textbf{Setting} & \textbf{B-4} & \textbf{M} & \textbf{R-L} &
\textbf{CIDEr} & \textbf{CE-F1} & \textbf{GREEN} \\
\midrule
Disabled  & \textbf{28.0} & \textbf{44.4} & \textbf{41.6} & 11.7 & 17.7 & \textbf{38.0} \\
Predicted & 27.9 & 44.3 & \textbf{41.6} & 12.3 & 20.4 & 36.3 \\
Reference & 27.9 & 44.3 & \textbf{41.6} &
\textbf{12.4} & \textbf{20.7} & 36.6 \\
\bottomrule
\end{tabular}
\normalsize
\caption{Fixed-checkpoint ICF diagnostic on RadGenome-ChestCT: Disabled removes ICF, Predicted uses predicted status, and Reference uses reference-report-derived status.}
\label{tab:ICF_modes}
\end{table}

With the checkpoint, visual inputs, and decoding settings fixed, Table~\ref{tab:ICF_modes} compares the three ICF settings. All values are multiplied by 100, and boldface marks the best result for each metric. Compared with Disabled, Predicted raises CIDEr from 11.7 to 12.3 and CE-F1 from 17.7 to 20.4, while GREEN drops from 38.0 to 36.3, B-4 and M each decrease by 0.1, and R-L is unchanged. This fixed-input comparison isolates the deployed ICF prompt beyond GPI and RPD, showing gains in abnormality coverage rather than overall clinical-semantic agreement.

Reference leaves B-4, M, and R-L unchanged but slightly improves CIDEr, CE-F1, and GREEN, raising CE-F1 from 20.4 to 20.7. Because regional difference evidence is unchanged, this contrast tests sensitivity to status quality. These small clinical-metric gains and unchanged text overlap show that better status estimates only modestly change the generated reports; Reference is therefore a targeted diagnostic rather than a universal performance upper bound.

% ===================== CONCLUSION =====================
\section{Conclusion}
We introduced ALTER for longitudinal 3D CT report generation, combining multimodal history, region-specific interval evidence, and change-aware prompting. Results on RadGenome-ChestCT and CTRG-Chest-548K demonstrate improved anatomically grounded reporting and cross-dataset transfer. Additional analyses further show that these improvements are broadly distributed across anatomical regions and are particularly evident in clinical-semantic quality. ALTER is intended to assist clinicians in longitudinal CT interpretation and report drafting, with future studies needed to further evaluate its performance in real-world clinical settings. We also plan to model the time elapsed between examinations.

\clearpage
\appendix

% ===================== Appendix A =====================
\section{Dataset Details}
\label{app:data}

\paragraph{Datasets.}
RadGenome-ChestCT is derived from CT-RATE and contains 25{,}692 non-contrast chest CT volumes from 21{,}304 patients. According to the released split, 24{,}128 volumes from 20{,}000 patients are used for training, and 1{,}564 volumes from 1{,}304 patients are used for validation. Each volume is paired with a corresponding radiology report. The dataset additionally provides segmentation masks covering 197 anatomical categories and approximately 665{,}000 region-grounded descriptions. We use RadGenome-ChestCT as the primary benchmark for region-aware report generation under different conditions of prior availability. CTRG-Chest-548K is an independent corpus comprising 1{,}804 chest CT volumes with Chinese-language source reports. Each model configuration is fine-tuned on the standard training split and evaluated on the 362-study test split. Because region-grounded descriptions are unavailable, the English report fields provided with the public split annotations are mapped to predefined anatomical regions as described below.

\paragraph{Split statistics.}
Table~\ref{tab:dataset_stats} summarizes the splits of each dataset according to the availability of an eligible prior. An eligible prior is defined as the immediately preceding CT volume from the same patient that remains available after all data-integrity checks. Prior availability is therefore determined by the released longitudinal sequence metadata rather than by a predefined sampling ratio. After the integrity checks, 24{,}126 of the 24{,}128 released RadGenome-ChestCT training studies remain, of which 4{,}127 have an eligible prior. The validation split contains 1{,}564 studies, including 260 with an eligible prior and 1{,}304 without one. Under the same selection criteria, none of the 362 CTRG-Chest-548K test studies has an eligible prior; this benchmark therefore evaluates report generation based solely on the current examination.

\paragraph{Prior handling.}
Each longitudinal sequence is reconstructed using the patient identifier and ordinal time point encoded in the released volume names. For each current study, only the immediately preceding examination in the sequence is considered. We set $h{=}1$ when this volume is available and $h{=}0$ otherwise, without substituting an earlier examination for a missing immediate predecessor. The corresponding prior report is retrieved using the preceding volume name. If the report is unavailable, the prior image remains eligible, while the textual history is represented as empty. When $h{=}1$, the history-dependent branches receive the preceding volume and its available report; when $h{=}0$, prior attention, regional differencing, and the ICF change pathway are disabled. This rule ensures ordinal adjacency within each released sequence but does not guarantee comparability in contrast phase, anatomical coverage, or elapsed time between examinations. Prior prevalence is therefore neither balanced nor experimentally controlled. We report both aggregate and prior-stratified results to distinguish history-conditioned generation from generation based solely on the current examination.

\begin{table}[t]
\centering
\footnotesize
\setlength{\tabcolsep}{1.5pt}
\begin{tabular}{@{}llrrrr@{}}
\toprule
\textbf{Dataset} & \textbf{Split} & \textbf{Total} &
\textbf{Prior} & \textbf{No prior} & \textbf{Rate} \\
\midrule
RadGenome-ChestCT & Train      & 24{,}126 & 4{,}127 & 19{,}999 & 17.1\% \\
RadGenome-ChestCT & Validation & 1{,}564  & 260     & 1{,}304  & 16.6\% \\
CTRG-Chest-548K   & Test       & 362      & 0       & 362      & 0\% \\
\bottomrule
\end{tabular}
\normalsize
\caption{Prior availability across the evaluated dataset splits.}
\label{tab:dataset_stats}
\end{table}

\begin{figure}[t]
\centering
\includegraphics[width=\columnwidth]{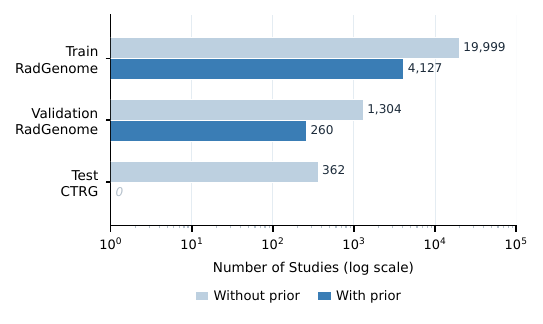}
\caption{Distribution of prior availability across the evaluated dataset splits.}
\label{fig:prior_dist}
\end{figure}

\subsection{CTRG Report Processing}
\label{app:ctrg_processing}

Although the source reports in CTRG-Chest-548K are written in Chinese, the public split annotations provide English findings and impressions for each study. We use these fields directly and normalize repeated whitespace before sentence segmentation. The findings and impressions are then concatenated, with sentence boundaries identified at periods and semicolons. The resulting sentences preserve the original wording of the provided English reports; subsequent processing modifies only their anatomical assignments.

Each report is submitted once to DeepSeek-v4-flash with the temperature set to zero. The prompt explicitly enumerates the 10 reporting regions and requests a compact mapping in JavaScript Object Notation that associates each numbered sentence with one or more exact region names. Sentences involving multiple regions are retained under all regions returned by the model. Invalid region names are removed, and sentences with no valid assignment default to the lung region. If the model response is unavailable or cannot be parsed, a deterministic keyword router assigns each sentence based on anatomy-specific terms, using the lung region only when no term is matched. This fallback is invoked for only one report in each split. The resulting assignments are cached by volume and stored as triples comprising the volume name, anatomical region, and original sentence.

Table~\ref{tab:ctrg_processing} summarizes the resulting region-level supervision. An assignment represents an association between one sentence and one region; therefore, a sentence mapped to two regions contributes two assignments. Mean regions denotes the number of distinct regions represented in each study. The similar statistics across the three splits indicate that the same processing procedure yields comparable anatomical coverage during training and evaluation.

\begin{table}[t]
\centering
\footnotesize
\setlength{\tabcolsep}{2.5pt}
\begin{tabular}{@{}lrrr@{}}
\toprule
\textbf{Split} & \textbf{Studies} & \textbf{Assignments} &
\textbf{Mean regions} \\
\midrule
Train      & 1{,}262 & 14{,}036 & 5.05 \\
Validation & 180     & 1{,}946  & 5.08 \\
Test       & 362     & 4{,}043  & 5.08 \\
\bottomrule
\end{tabular}
\normalsize
\caption{CTRG report-processing statistics.}
\label{tab:ctrg_processing}
\end{table}

Table~\ref{tab:ctrg_example} illustrates the resulting assignments for one test report. The heart and mediastinum sentence demonstrates that the mapping is not restricted to a single region, while the pleural and adrenal findings are assigned to their corresponding anatomical contexts.

\begin{table}[t]
\centering
\footnotesize
\setlength{\tabcolsep}{3pt}
\begin{tabular}{@{}p{0.66\columnwidth}p{0.26\columnwidth}@{}}
\toprule
\textbf{Report sentence} & \textbf{Region} \\
\midrule
The shape of heart shadow and large vessels is normal, and no mass or enlarged lymph node is found in the mediastinum. & Heart; \mbox{mediastinum} \\
There is a small amount of pleural effusion on the right side, and the pleural effusion on the left side is slightly thicker. & Pleura \\
A low-density nodule is seen in the left adrenal gland. & Abdomen \\
\bottomrule
\end{tabular}
\normalsize
\caption{Example CTRG report assignments.}
\label{tab:ctrg_example}
\end{table}

% ===================== Appendix B =====================
\section{Experimental Setup}
\label{app:impl}

\subsection{Evaluation Protocol}

\paragraph{Evaluation settings.}
ALTER is evaluated on the RadGenome-ChestCT validation split comprising 1{,}564 studies. We report both aggregate results and results stratified by prior availability. On CTRG-Chest-548K, Reg2RG, DEAR, and ALTER are fine-tuned on the training set and evaluated on the 362-study test split. Under our prior-selection protocol, none of the test studies has an eligible prior; this benchmark therefore evaluates report generation based solely on the current examination. For R2GenGPT, MedVInT, RadFM, CT2Rep, and M3D, we report previously published results; their CTRG-Chest-548K scores were obtained on a different split of this dataset.

\paragraph{CTRG adaptation.}
CTRG-Chest-548K does not provide the anatomical segmentation masks required by the regional reporting interface. We therefore apply TotalSegmentator to each volume and merge the available anatomical structures into the same reporting regions used for RadGenome-ChestCT. The pleural region is approximated using the outer shell of the lung masks, while regions without valid masks are excluded from regional supervision. Report targets are constructed following the procedure described in Section~\ref{app:ctrg_processing}. The same visual and textual processing procedures are applied to the training, validation, and test splits to ensure consistency across data partitions.

\subsection{Model Configuration}

Across all experiments, TotalSegmentator masks are aggregated into $R{=}10$ anatomical reporting regions. A Vision Transformer initialized from the RadFM 3D visual checkpoint is kept frozen and used to encode both full volumes and regional inputs. A shared Perceiver resampler compresses each feature sequence into 32 latent tokens, which are subsequently projected into the 4{,}096-dimensional embedding space of RadLLaMA-7B. The report decoder follows the radiology-adapted Reg2RG configuration. Prior reports are truncated to a maximum of 512 tokens and encoded by the frozen RadBERT-RoBERTa-4m model, whose outputs are projected into the same embedding space. The decoder is adapted using low-rank adaptation with rank 8 and $\alpha{=}32$.

\paragraph{Optimization.}
The configuration comprising Global Prior Integration (GPI) and RPD is trained for 10 epochs using AdamW, with a peak learning rate of $5\!\times\!10^{-5}$ and 100 linear warmup steps. This stage uses three graphics processing units (GPUs) with four gradient-accumulation steps. ICF fine-tuning is initialized from the selected GPI and RPD checkpoint and uses two GPUs with six gradient-accumulation steps. DeepSpeed state sharding is not enabled. Difference aggregation, soft-prompt projection layers, and their associated normalization layers operate in 32-bit floating-point precision, while all remaining operations use bfloat16 (BF16) mixed precision. The difference-projection weights are initialized with a standard deviation of 0.01. The ICF status and change gates are initialized to $\gamma_s{=}0.02$ and $\gamma_c{=}0.01$, respectively. These settings are held fixed across all reported configurations.

\paragraph{Supervision.}
ICF uses four soft-prompt tokens. The status head receives gradient-detached features and is optimized using an auxiliary binary classification loss with a weight of 0.1. Abnormality pseudo-labels derived from the reference reports supervise the status prediction under an 18-class abnormality ontology.

\subsection{Computational Cost}

\begin{figure}[t]
\centering
\includegraphics[width=\columnwidth]{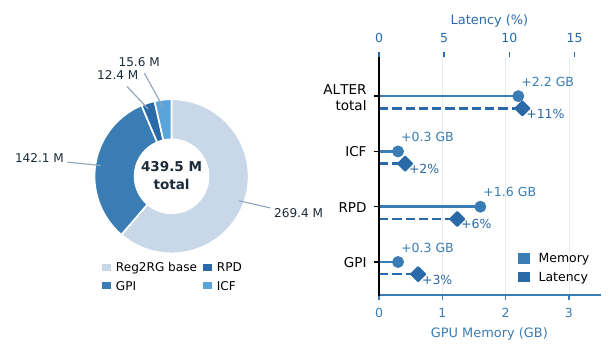}
\caption{Computational profile of ALTER.}
\label{fig:compute_cost}
\end{figure}

Table~\ref{tab:compute_cost} summarizes the number of trainable parameters and the per-study inference overhead. The Reg2RG and ALTER rows report the total numbers of trainable parameters, whereas the module-specific rows report parameter increments relative to the base model. ALTER contains 439.5 million trainable parameters, representing an increase of 170.1 million over Reg2RG. These counts exclude the parameters of the frozen encoders and those of the language model outside the adapters. On a single NVIDIA RTX PRO 6000 Blackwell GPU with a batch size of 1 and BF16 precision, the measured RPD and ICF increments jointly add 1.9 GB of GPU memory and 8\% inference latency. GPI is not profiled independently because it shares the prior-volume encoding pass with the full model. Its 0.3 GB memory increment is calculated from its parameter-storage requirement, while its 3\% latency increment is estimated. Accordingly, the ALTER row aggregates the measured and estimated module-level quantities rather than reporting a separate end-to-end measurement. Figure~\ref{fig:compute_cost} visualizes the same computational profile. The left panel decomposes the trainable parameter counts and shows that GPI contributes the largest increment, whereas the right panel presents the memory and latency increments of each module and their aggregate.

\begin{table}[t]
\centering
\footnotesize
\setlength{\tabcolsep}{2pt}
\begin{tabular}{@{}lccc@{}}
\toprule
\textbf{Model/Module} & \textbf{Parameters (M)} &
\textbf{Memory ($\Delta$)} & \textbf{Latency ($\Delta$)} \\
\midrule
Reg2RG & 269.4 & -- & -- \\
GPI & 142.1 & $+$0.3 GB & $+$3\% \\
RPD & 12.4 & $+$1.6 GB & $+$6\% \\
ICF & 15.6 & $+$0.3 GB & $+$2\% \\
\midrule
\textbf{ALTER} & \textbf{439.5} & $+$2.2 GB & $+$11\% \\
\bottomrule
\end{tabular}
\normalsize
\caption{Trainable parameters and inference overhead relative to Reg2RG.}
\label{tab:compute_cost}
\end{table}

% ===================== Appendix C =====================
\section{Additional Results}
\label{app:results}

The supplementary experiments further evaluate the design of the RPD proxy and its performance across anatomical regions. These analyses assess the effectiveness of region-specific historical retrieval and characterize the anatomical distribution of the resulting performance gains.

\subsection{RPD Proxy}
\label{app:ablation}

\begin{figure}[t]
\centering
\includegraphics[width=\columnwidth]{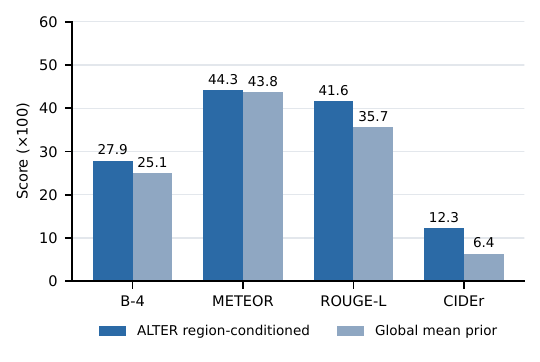}
\caption{Effect of RPD proxy design on RadGenome-ChestCT.}
\label{fig:rpd_proxy}
\end{figure}

Table~\ref{tab:ablation_dcc} and Figure~\ref{fig:rpd_proxy} compare the region-conditioned RPD proxy with a global-mean alternative. This alternative replaces region-wise cross-attention with mean pooling over the prior tokens and broadcasts the resulting single representation to all current regions. All other training and inference settings are held constant. All values are multiplied by 100, and boldface indicates the better result in each table column. Compared with the global-mean alternative, region-conditioned retrieval raises B-4 from 25.1 to 27.9, R-L from 35.7 to 41.6, M from 43.8 to 44.3, and CIDEr from 6.4 to 12.3. R-L and CIDEr show the largest absolute gains of 5.9 points each, while the other two metrics also improve. These results suggest that compressing the prior information into a single global representation may discard historical information specific to individual anatomical regions. In contrast, conditioning retrieval on each current region produces region-differentiated historical proxies, thereby providing more targeted longitudinal evidence for report generation.

\begin{table}[t]
\centering
\small
\begin{tabular}{@{}lcccc@{}}
\toprule
\textbf{Configuration} & \textbf{B-4} & \textbf{R-L} & \textbf{M} & \textbf{CIDEr} \\
\midrule
ALTER            & \textbf{27.9} & \textbf{41.6} & \textbf{44.3} & \textbf{12.3} \\
Global mean prior       & 25.1 & 35.7 & 43.8 & 6.4 \\
\bottomrule
\end{tabular}
\normalsize
\caption{Comparison of RPD proxy designs on RadGenome-ChestCT.}
\label{tab:ablation_dcc}
\end{table}

\subsection{Abnormality Results}
\label{app:abnormality_results}

Table~\ref{tab:abnormality_ce} further decomposes the aggregate clinical efficacy (CE) results reported in the main paper into the 18 abnormality categories defined by the fixed clinical labeler. Both methods are evaluated on the same RadGenome-ChestCT studies using the same reference reports. Support denotes the number of positive reference labels for each category, all metric values are multiplied by 100, and boldface indicates the higher F1 score for each abnormality.

ALTER raises weighted CE precision from 49.4 to 49.6, while recall decreases from 16.1 to 12.8 and F1 from 24.3 to 20.4. It obtains a higher class-specific F1 for 3 of the 18 abnormalities. The largest gains occur for lung opacity, coronary artery wall calcification, and arterial wall calcification, which improve by 8.9, 4.0, and 3.6 points, respectively. This pattern shows that the aggregate CE decrease is spread across multiple pulmonary, pleural, and cardiac findings rather than driven primarily by a single dominant category.

The class-wise results also exhibit substantial nonuniformity. Reg2RG retains higher F1 scores on the remaining 15 abnormalities, while ALTER records zero F1 for medical material, pericardial effusion, hiatal hernia, peribronchial thickening, and bronchiectasis. Both methods yield low recall on several abnormalities, especially pericardial effusion, hiatal hernia, peribronchial thickening, and lymphadenopathy. The difference in weighted CE metrics therefore reflects a more conservative precision--recall balance rather than an improvement in recognition quality.

\begin{table*}[t]
\centering
\setlength{\tabcolsep}{5pt}
\begin{tabular}{@{}lrrrrrrr@{}}
\toprule
& & \multicolumn{3}{c}{\textbf{Reg2RG}} & \multicolumn{3}{c}{\textbf{ALTER}} \\
\cmidrule(lr){3-5}\cmidrule(lr){6-8}
\textbf{Abnormality} & \textbf{Support} & \textbf{P} & \textbf{R} & \textbf{F1} & \textbf{P} & \textbf{R} & \textbf{F1} \\
\midrule
Medical material                    & 139 & 13.5 & 9.4 & \textbf{11.1} & 0.0 & 0.0 & 0.0 \\
Arterial wall calcification         & 412 & 72.4 & 13.3 & 22.5 & 74.7 & 15.8 & \textbf{26.1} \\
Cardiomegaly                        & 159 & 43.3 & 36.5 & \textbf{39.6} & 43.4 & 22.6 & 29.8 \\
Pericardial effusion                & 103 & 28.6 & 1.9 & \textbf{3.6} & 0.0 & 0.0 & 0.0 \\
Coronary artery wall calcification  & 351 & 70.0 & 14.0 & 23.3 & 67.4 & 17.1 & \textbf{27.3} \\
Hiatal hernia                       & 197 & 100.0 & 0.5 & \textbf{1.0} & 0.0 & 0.0 & 0.0 \\
Lymphadenopathy                     & 337 & 31.8 & 2.1 & \textbf{3.9} & 100.0 & 0.3 & 0.6 \\
Emphysema                           & 302 & 41.3 & 17.2 & \textbf{24.3} & 42.3 & 10.9 & 17.4 \\
Atelectasis                         & 356 & 36.2 & 14.0 & \textbf{20.2} & 40.0 & 6.2 & 10.7 \\
Lung nodule                         & 681 & 46.9 & 22.5 & \textbf{30.4} & 51.6 & 14.4 & 22.5 \\
Lung opacity                        & 604 & 72.0 & 29.8 & 42.2 & 72.6 & 39.4 & \textbf{51.1} \\
Pulmonary fibrotic sequela          & 405 & 37.5 & 8.9 & \textbf{14.4} & 44.6 & 6.2 & 10.8 \\
Pleural effusion                    & 164 & 63.3 & 30.5 & \textbf{41.2} & 77.1 & 16.5 & 27.1 \\
Mosaic attenuation pattern          & 125 & 32.9 & 21.6 & \textbf{26.1} & 27.0 & 8.0 & 12.3 \\
Peribronchial thickening            & 161 & 12.5 & 0.6 & \textbf{1.2} & 0.0 & 0.0 & 0.0 \\
Consolidation                       & 275 & 47.7 & 22.2 & \textbf{30.3} & 46.8 & 10.5 & 17.2 \\
Bronchiectasis                      & 160 & 18.3 & 8.1 & \textbf{11.3} & 0.0 & 0.0 & 0.0 \\
Interlobular septal thickening      & 117 & 36.8 & 6.0 & \textbf{10.3} & 22.2 & 3.4 & 5.9 \\
\midrule
Weighted average                    & 5{,}048 & 49.4 & 16.1 & \textbf{24.3} & 49.6 & 12.8 & 20.4 \\
\bottomrule
\end{tabular}
\normalsize
\caption{Abnormality-level clinical efficacy on RadGenome-ChestCT.}
\label{tab:abnormality_ce}
\end{table*}

\subsection{Regional Results}
\label{app:region_ce}

\begin{figure*}[t]
\centering
\includegraphics[width=\textwidth]{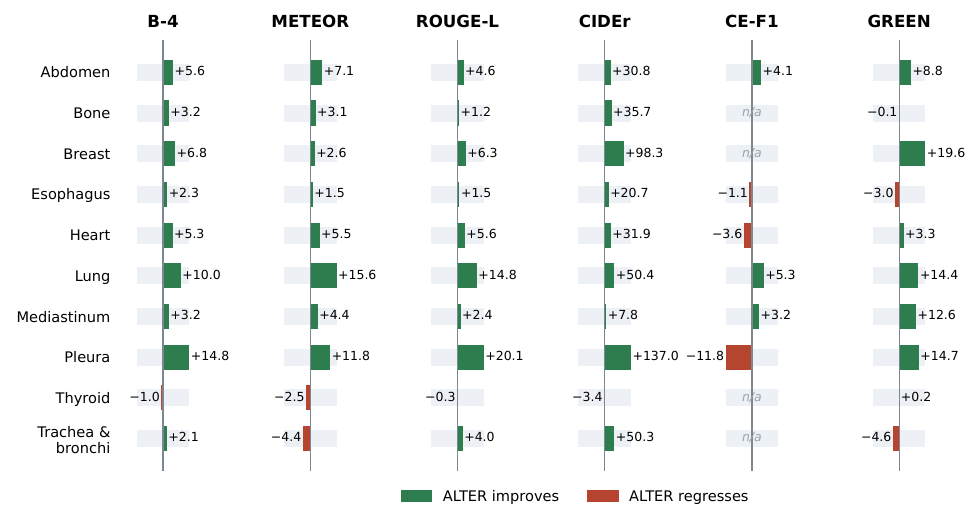}
\caption{Per-region score differences between ALTER and Reg2RG.}
\label{fig:region_delta}
\end{figure*}

Table~\ref{tab:region_nlg} and Figure~\ref{fig:region_delta} compare ALTER and Reg2RG across the 10 reporting regions defined in RadGenome-ChestCT. Evaluation is conducted using the deployed \textit{Predicted} mode and the protocol adopted in the main experiments. Each regional score is computed only from studies containing a reference description for the corresponding region. The value $n$ denotes the number of such studies, and all metric values are multiplied by 100. In the table, CE-F1 and GREEN are abbreviated as CE and GRN, respectively. Reference descriptions appear in only 56 validation studies for the breast region and 51 for the thyroid region; estimates for these regions should therefore be interpreted with particular caution. The Overall row reports study-level performance rather than the average of the regional scores. Figure~\ref{fig:region_delta} shows the performance difference between ALTER and Reg2RG, computed as ALTER minus Reg2RG, such that positive values favor ALTER.

\paragraph{Text-overlap performance.}
ALTER improves all four text-overlap metrics in 8 of the 10 regions. The three largest CIDEr gains occur in the pleura, breast, and lung regions at 137.0, 98.3, and 50.4 points, respectively, though the breast gain is based on limited support. The remaining two regions show metric-dependent patterns. For the trachea and bronchi, B-4, R-L, and CIDEr increase, while M decreases by 4.4 points. In the thyroid region, ALTER produces lower B-4, M, R-L, and CIDEr scores, but limited support prevents a reliable comparison. Overall, the regional text-performance gains extend across diverse anatomical structures, although their magnitude and consistency across metrics vary by region.

\paragraph{Clinical performance.}
CE-F1 is available only for the six regions represented in the 18-class RadBERT label space; dashes indicate the remaining four regions. ALTER raises CE-F1 for the lung, abdomen, and mediastinum by 5.3, 4.1, and 3.2 points, respectively, but yields lower scores for the pleura, heart, and esophagus. GREEN increases in 7 of the 10 regions, with the largest gains in the breast, pleura, and lung at 19.6, 14.7, and 14.4 points, respectively. Decreases occur only in the bone, esophagus, and trachea and bronchi regions. Overall, ALTER improves text overlap and clinical-semantic quality across multiple anatomical regions.

\begin{table*}[t]
\centering
\setlength{\tabcolsep}{3pt}
\begin{tabular}{@{}lccccccccccccc@{}}
\toprule
 & & \multicolumn{6}{c}{\textbf{Reg2RG}} & \multicolumn{6}{c}{\textbf{ALTER}} \\
\cmidrule(lr){3-8}\cmidrule(lr){9-14}
\textbf{Region} & \textbf{n} & \textbf{B-4} & \textbf{M} & \textbf{R-L} & \textbf{CIDEr} & \textbf{CE} & \textbf{GRN} & \textbf{B-4} & \textbf{M} & \textbf{R-L} & \textbf{CIDEr} & \textbf{CE} & \textbf{GRN} \\
\midrule
Abdomen             & 1{,}518 & 25.9 & 42.8 & 39.8 & 114.9 & 18.0 & 46.6 & 31.5 & 49.9 & 44.4 & 145.7 & 22.1 & 55.4 \\
Bone                & 1{,}509 & 30.9 & 45.9 & 43.8 & 216.6 & -- & 38.1 & 34.1 & 49.0 & 45.0 & 252.3 & -- & 38.0 \\
Breast              & 56 & 13.9 & 24.6 & 23.2 & 57.6 & -- & 9.0 & 20.7 & 27.2 & 29.5 & 155.9 & -- & 28.6 \\
Esophagus           & 1{,}324 & 53.4 & 63.3 & 64.3 & 353.1 & 1.1 & 82.6 & 55.7 & 64.8 & 65.8 & 373.8 & 0.0 & 79.6 \\
Heart               & 1{,}422 & 28.1 & 44.6 & 43.4 & 170.6 & 19.0 & 59.1 & 33.4 & 50.1 & 49.0 & 202.5 & 15.4 & 62.4 \\
Lung                & 1{,}514 & 8.3 & 18.9 & 20.0 & 15.6 & 14.7 & 12.7 & 18.3 & 34.5 & 34.8 & 66.0 & 20.0 & 27.1 \\
Mediastinum         & 1{,}516 & 18.5 & 37.8 & 32.7 & 56.2 & 15.3 & 50.3 & 21.7 & 42.2 & 35.1 & 64.0 & 18.5 & 62.9 \\
Pleura              & 1{,}172 & 15.7 & 27.9 & 30.1 & 62.3 & 17.8 & 20.3 & 30.5 & 39.7 & 50.2 & 199.3 & 6.0 & 35.0 \\
Thyroid             & 51 & 10.4 & 25.9 & 26.2 & 13.6 & -- & 20.3 & 9.4 & 23.4 & 25.9 & 10.2 & -- & 20.5 \\
Trachea and bronchi & 1{,}412 & 39.9 & 60.0 & 57.6 & 244.1 & -- & 82.7 & 42.0 & 55.6 & 61.6 & 294.4 & -- & 78.1 \\
\midrule
Overall              & 1{,}564 & 25.2 & 42.7 & 37.2 & 7.3 & 24.3 & 33.7 & 27.9 & 44.3 & 41.6 & 12.3 & 20.4 & 36.3 \\
\bottomrule
\end{tabular}
\normalsize
\caption{Per-region performance on RadGenome-ChestCT.}
\label{tab:region_nlg}
\end{table*}

Taken together, the regional results show that ALTER's gains extend across diverse anatomical structures rather than being driven by a small subset of regions. The broadly distributed improvements in text-overlap and GREEN scores support the effectiveness of anatomy-aware longitudinal modeling for region-specific report generation, while CE-F1 captures complementary variation in clinical-entity recognition across regions.

% ===================== Appendix D =====================
\section{Longitudinal Cases}
\label{app:qual}

\begin{figure*}[!t]
\centering
\newcommand{\hit}[1]{{\sethlcolor{hlgreen}\hl{#1}}}
\newcommand{\halu}[1]{{\sethlcolor{hlred}\hl{#1}}}
\small
\setlength{\tabcolsep}{2pt}
\begin{tabular}{@{}p{0.175\textwidth}@{\hspace{4pt}}p{0.235\textwidth}@{\hspace{4pt}}p{0.235\textwidth}@{\hspace{4pt}}p{0.235\textwidth}@{}}
\toprule
\multicolumn{1}{c}{\textbf{CT Pair}} &
\multicolumn{1}{c}{\textbf{Reference}} &
\multicolumn{1}{c}{\textbf{Reg2RG}} &
\multicolumn{1}{c}{\textbf{ALTER}} \\
\midrule
\raisebox{\dimexpr-\height+\ht\strutbox\relax}{%
\begin{tabular}{@{}cc@{}}
\includegraphics[width=0.081\textwidth]{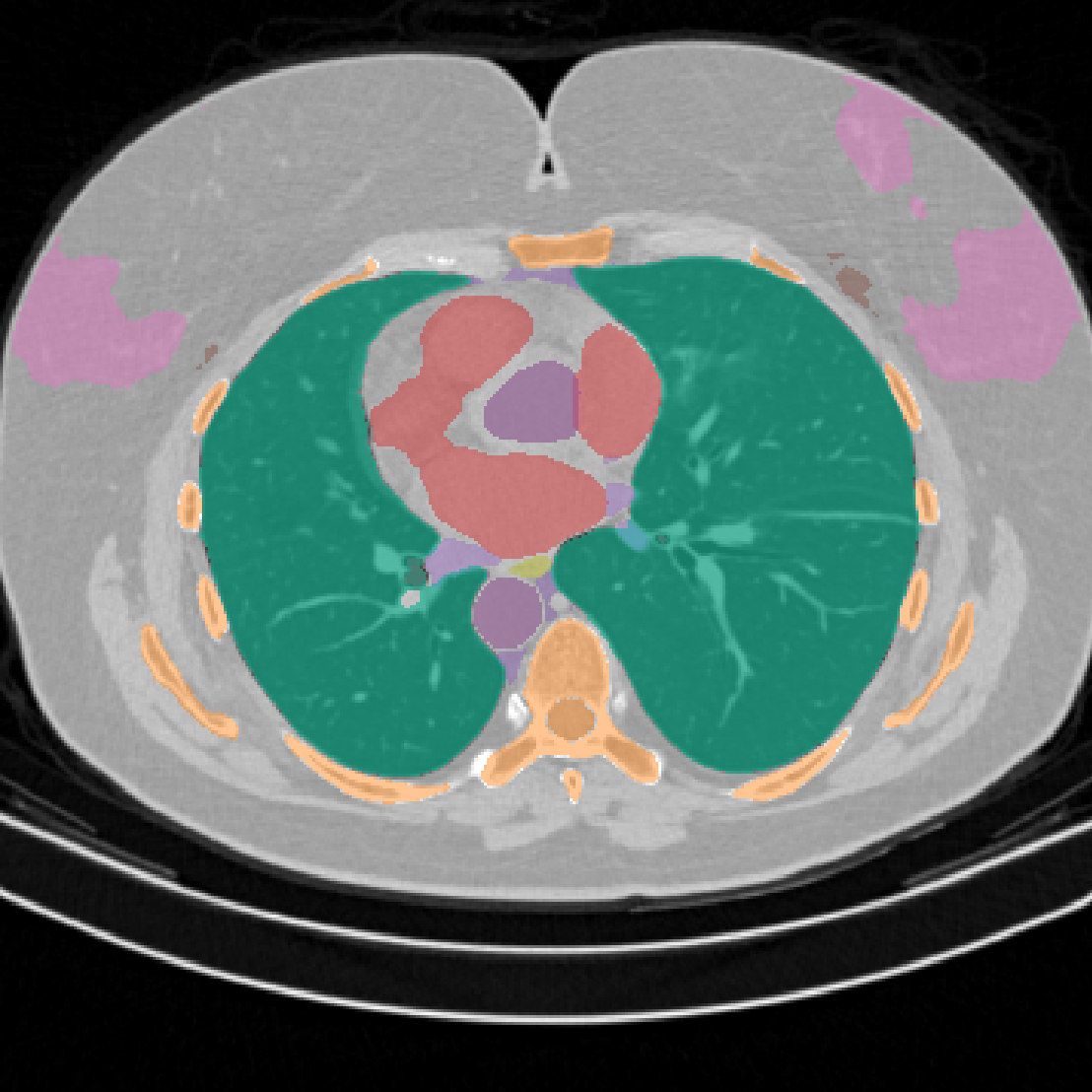} &
\includegraphics[width=0.081\textwidth]{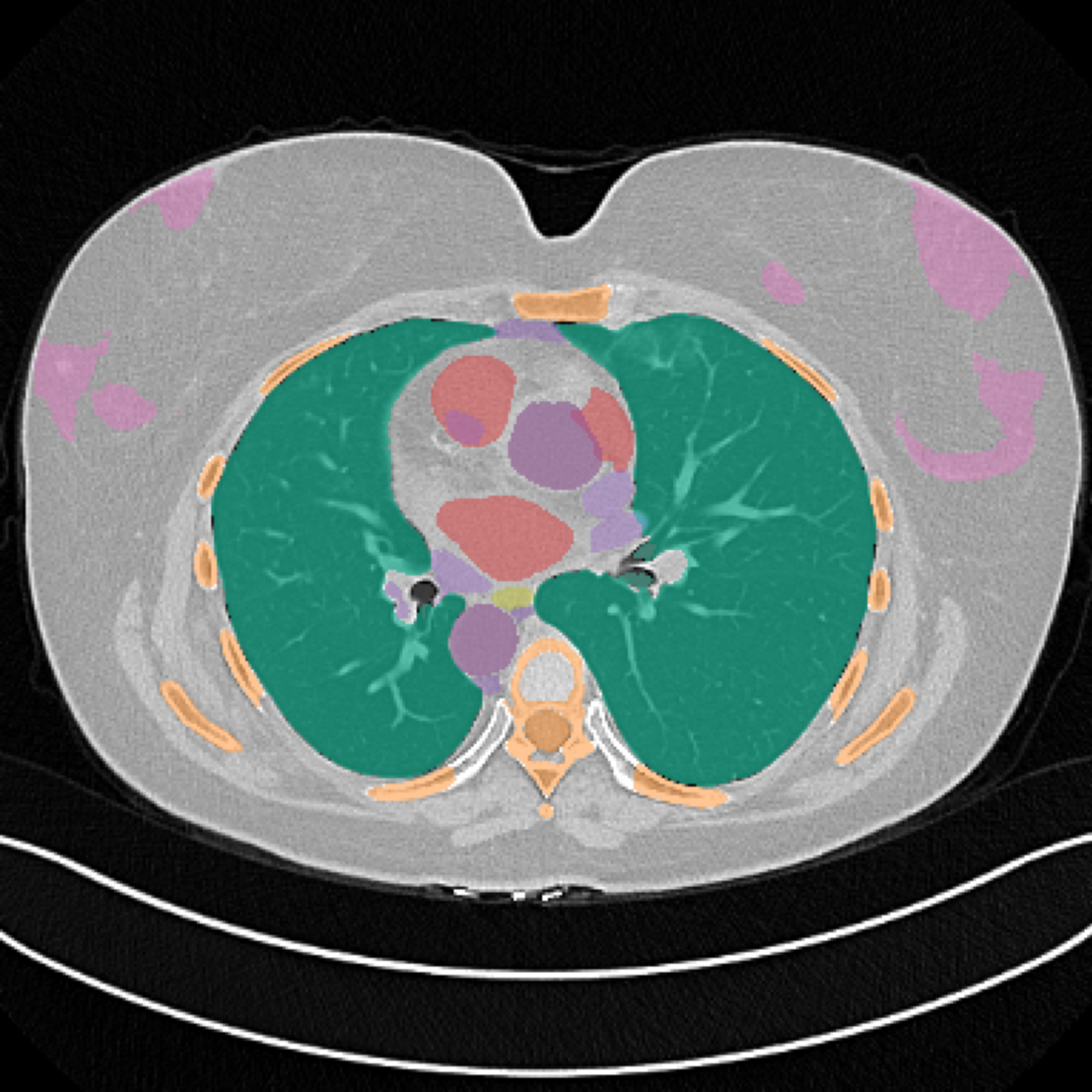} \\
{\scriptsize Prior} & {\scriptsize Current}
\end{tabular}} &
A millimetric nonspecific nodule is observed in the right lung and there is no significant difference. &
There are \halu{several millimetric nonspecific nodules in both lungs}. &
There is a \hit{stable nodule} in the right lung upper lobe posterior segment, which is 3 mm in diameter. \\
\midrule
\raisebox{\dimexpr-\height+\ht\strutbox\relax}{%
\begin{tabular}{@{}cc@{}}
\includegraphics[width=0.081\textwidth]{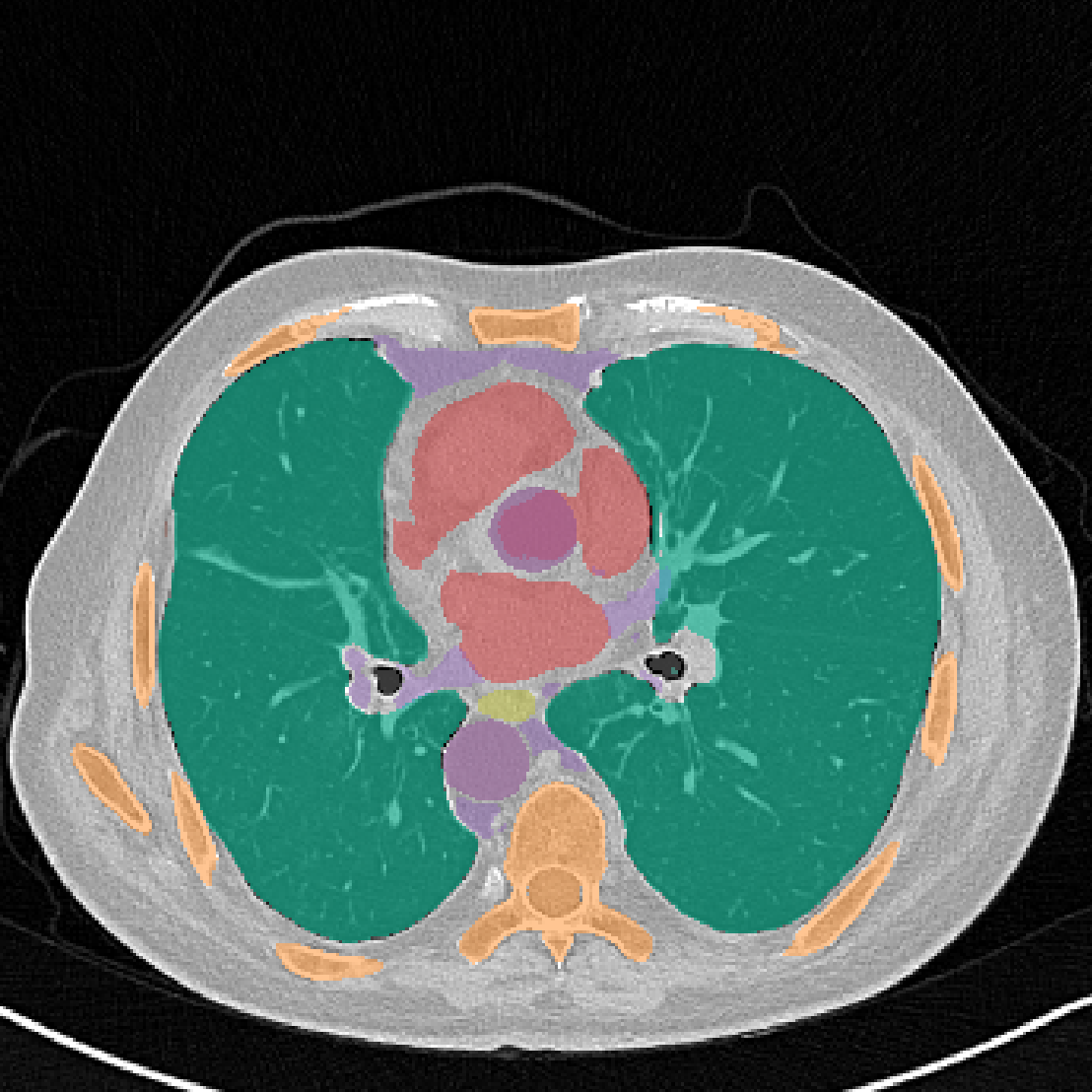} &
\includegraphics[width=0.081\textwidth]{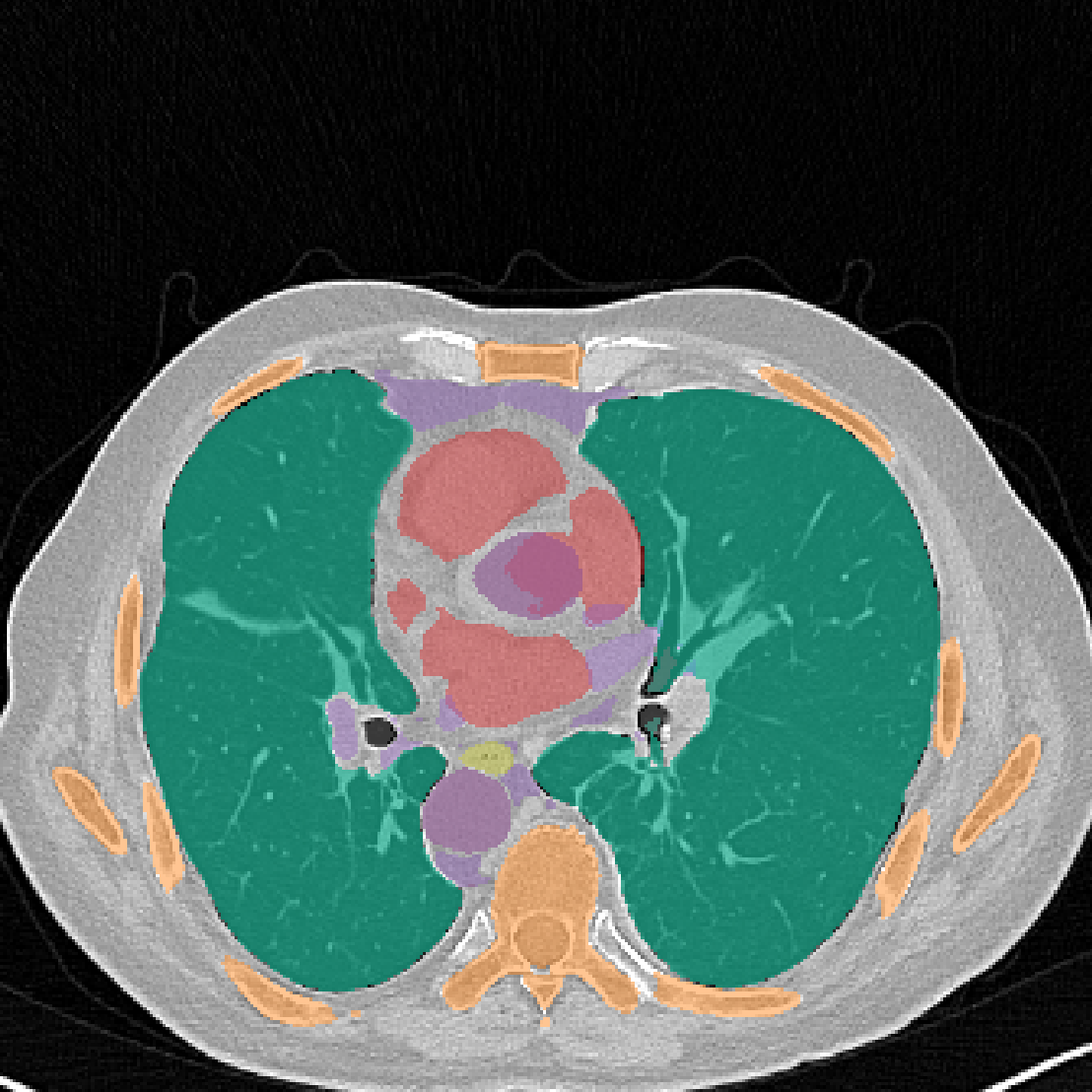} \\
{\scriptsize Prior} & {\scriptsize Current}
\end{tabular}} &
In both lungs, there are areas of increase in density consistent with newly developed consolidation, which is evaluated in favor of compressive atelectasis adjacent to the effusion. &
There are linear atelectasis in the lower lobe of the left lung and upper lobe of the right lung. \halu{No mass or infiltrative lesion was detected in both lungs}. &
In the current examination, there are \hit{newly developed} ground glass densities in the middle lobe of the right lung and in the anterior segments of both upper lobes. \\
\midrule
\raisebox{\dimexpr-\height+\ht\strutbox\relax}{%
\begin{tabular}{@{}cc@{}}
\includegraphics[width=0.081\textwidth]{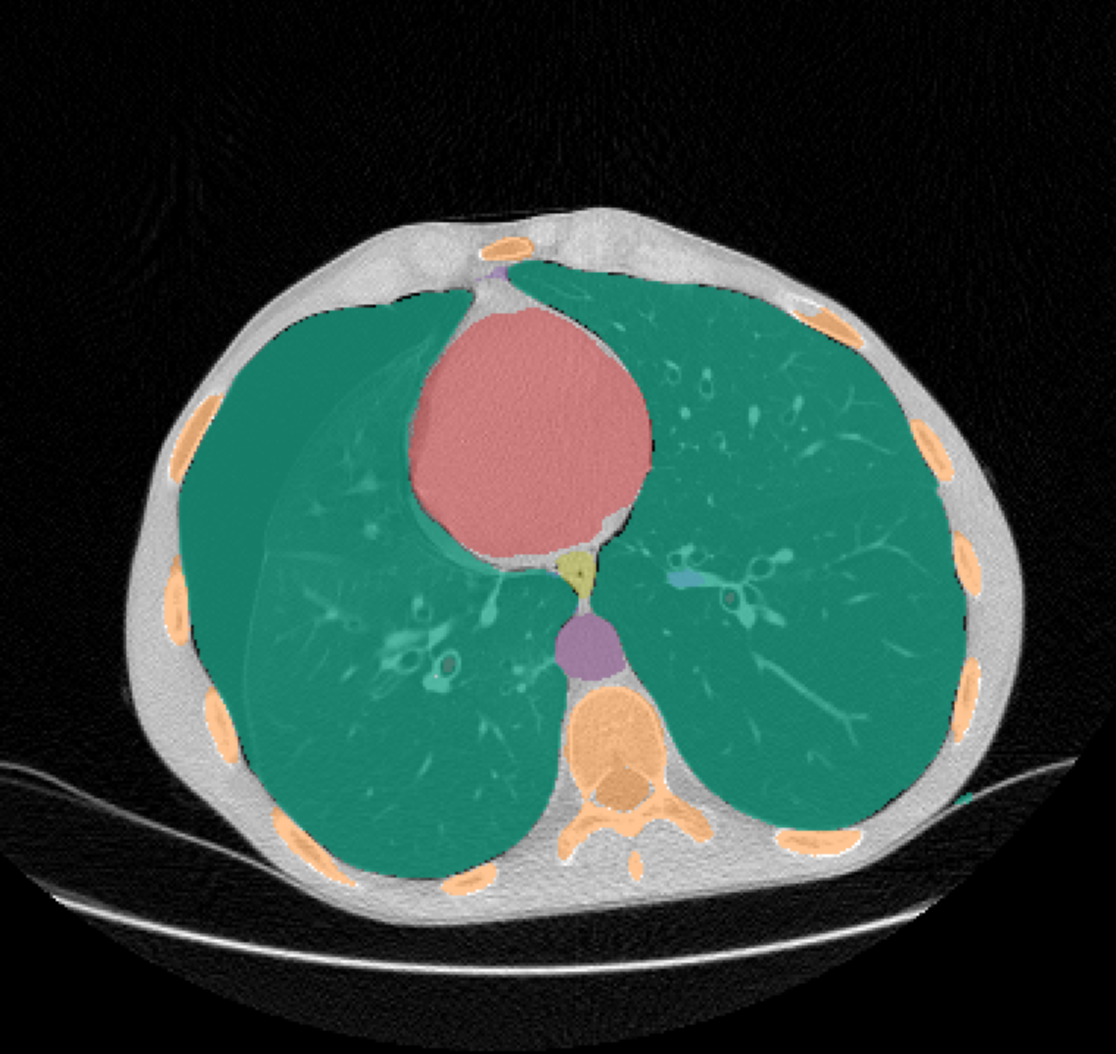} &
\includegraphics[width=0.081\textwidth]{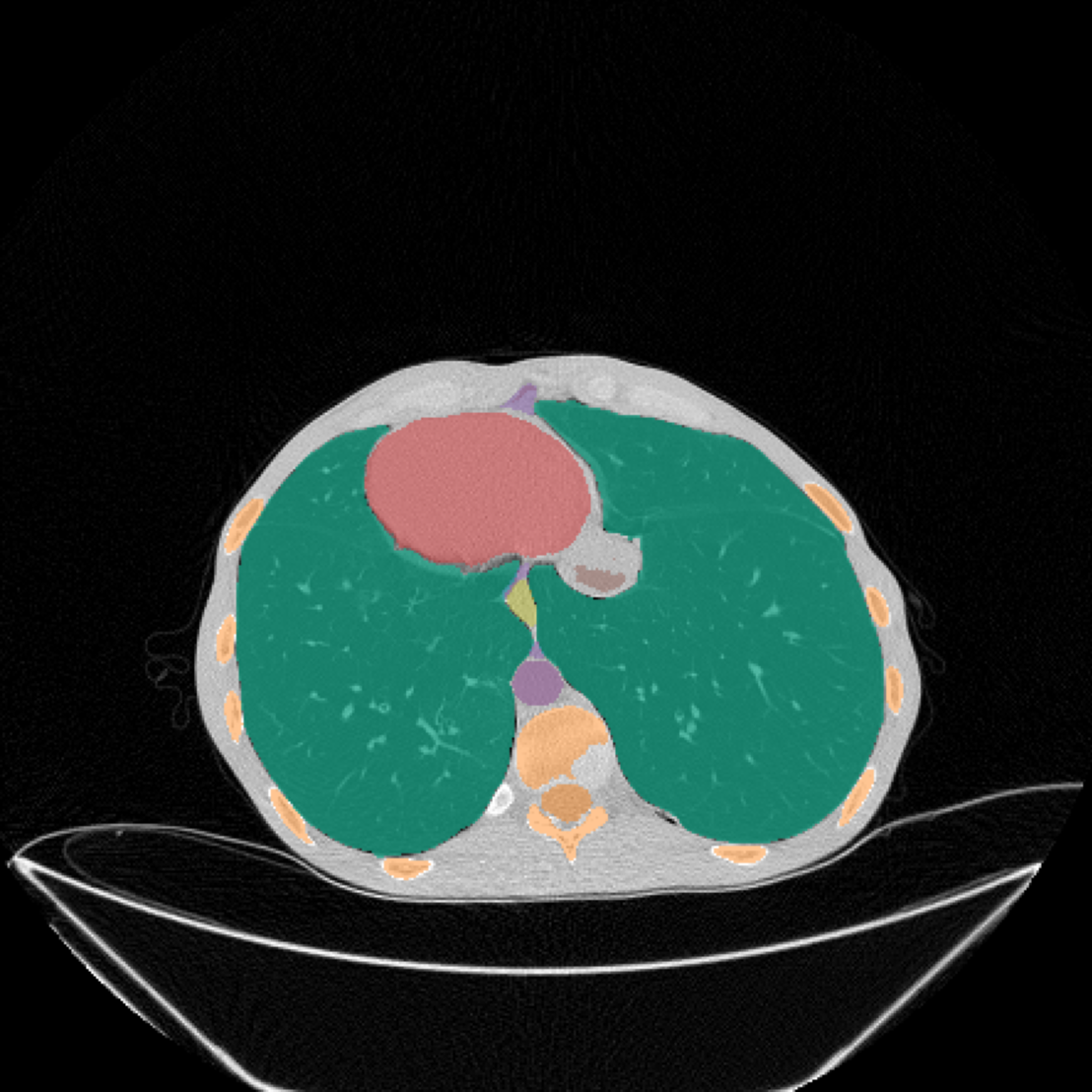} \\
{\scriptsize Prior} & {\scriptsize Current}
\end{tabular}} &
The left pneumothorax regressed significantly, while consolidative density progressed and a right pneumothorax newly emerged. &
Aeration of both lung parenchyma is normal and \halu{no nodular or infiltrative lesion is detected} in the lung parenchyma. &
Ground glass densities and consolidation areas in the right lung were observed in the previous examination and \hit{have increased in the current examination}. \\
\midrule
\raisebox{\dimexpr-\height+\ht\strutbox\relax}{%
\begin{tabular}{@{}cc@{}}
\includegraphics[width=0.081\textwidth]{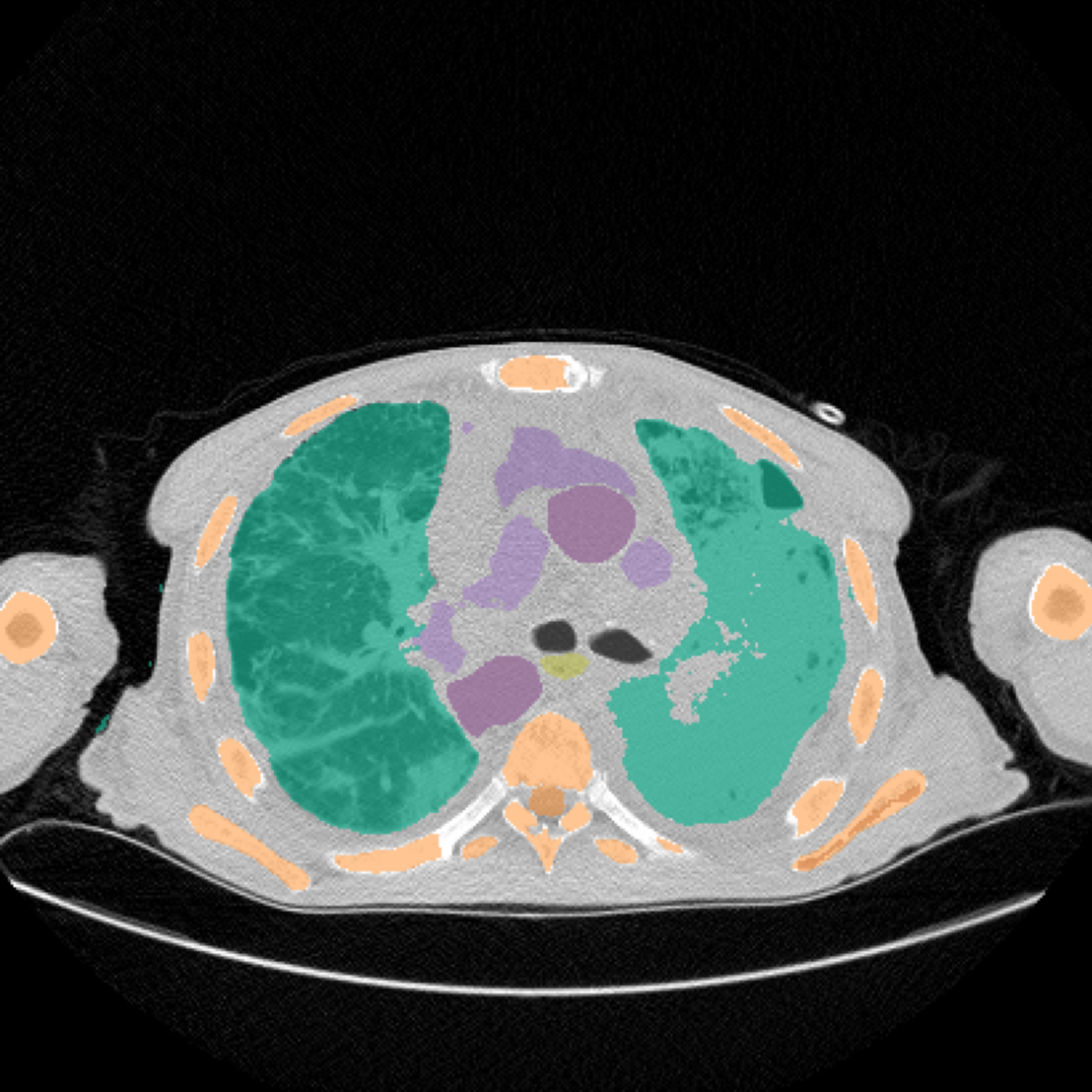} &
\includegraphics[width=0.081\textwidth]{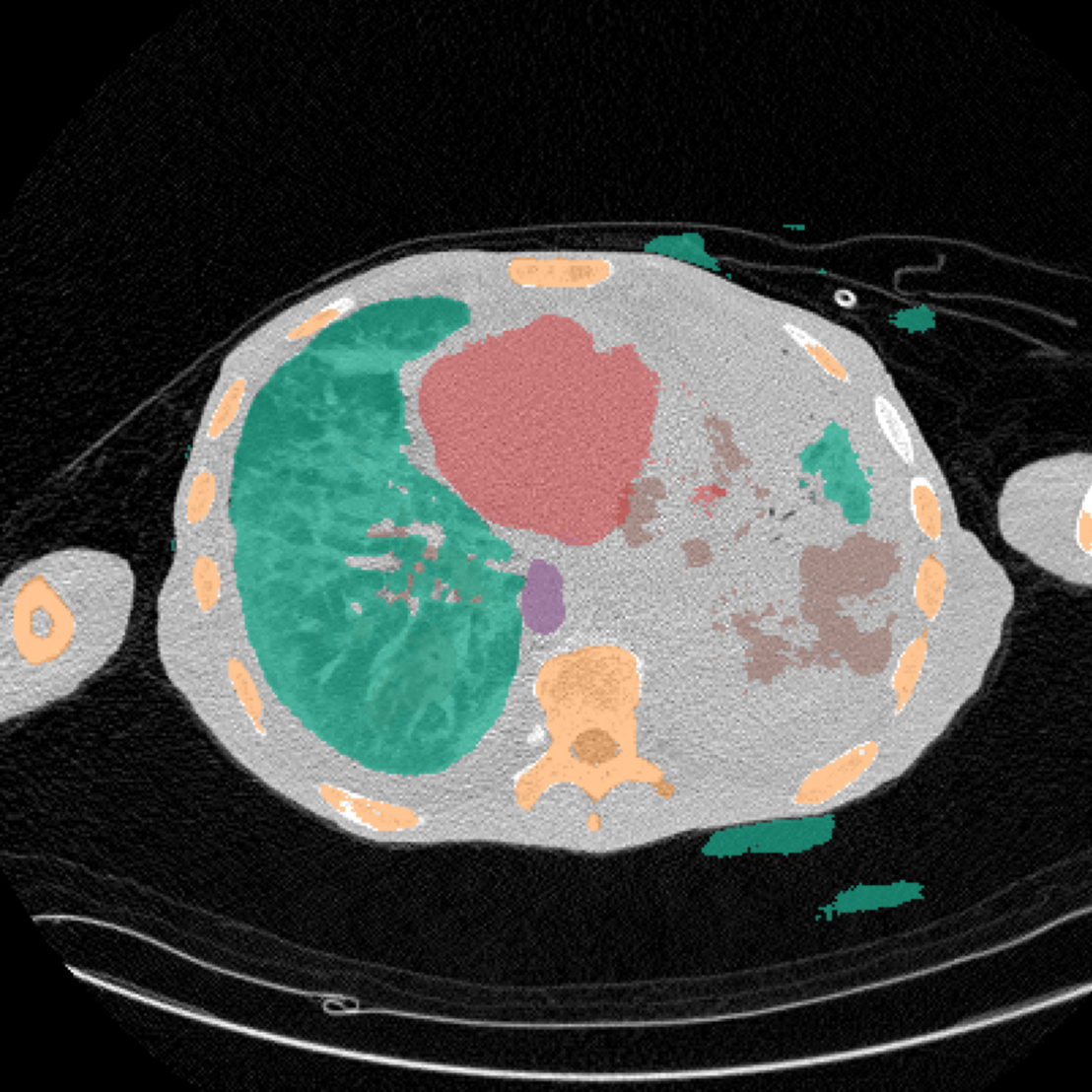} \\
{\scriptsize Prior} & {\scriptsize Current}
\end{tabular}} &
Widespread abdominal free fluid has just emerged. Liver and bone metastases are present, and the left upper-lobe soft-tissue density shows significant regression. &
\halu{Upper abdominal organs included in the sections are normal}. \halu{Bone structures in the study area are natural}. &
\halu{No solid mass was detected in the upper abdomen}. A right lower-lobe consolidation was \halu{evaluated in favor of regression of the infectious process}. \\
\bottomrule
\end{tabular}
\normalsize
\caption{Longitudinal comparison across four RadGenome-ChestCT cases.}
\label{fig:qualitative}
\end{figure*}

Figure~\ref{fig:qualitative} qualitatively compares ALTER and Reg2RG in four RadGenome-ChestCT validation studies with an eligible prior examination. The first column presents paired current and prior axial views with colored anatomical overlays, followed by temporally relevant excerpts from the reference and generated reports. In the ALTER column, green highlighting indicates content supported by the reference report, whereas red highlighting indicates unsupported or contradictory content.

Case 1 presents a stable pulmonary nodule. Reg2RG overstates both the number and bilateral distribution of the nodules and does not describe their interval status, whereas ALTER explicitly identifies the right lung nodule as stable. Case 2 contains newly developed bilateral pulmonary opacities adjacent to pleural effusions. ALTER correctly recognizes their new appearance but describes them as ground-glass densities rather than the consolidation reported in the reference, while Reg2RG states that no infiltrative lesion is present. Case 3 exhibits heterogeneous temporal changes, including increased consolidation, regression of a left pneumothorax, and a newly developed right pneumothorax. ALTER captures the increase in consolidation but does not describe either pneumothorax trajectory, whereas Reg2RG characterizes the lungs as normal. Case 4 presents concurrent abdominal, osseous, and pulmonary changes. Both methods omit the newly developed abdominal fluid and metastatic findings. In addition, ALTER incorrectly attributes regression to a right lower-lobe consolidation rather than the left upper-lobe lesion described in the reference.

Across the first three cases, historical conditioning enables ALTER to explicitly describe stability, new appearance, and interval increase, none of which are reflected in the Reg2RG outputs. However, the fourth case exposes a remaining limitation of ALTER: it misses the newly developed abdominal fluid and metastatic findings and incorrectly attributes regression to a right lower-lobe consolidation. This failure highlights the need to preserve the correspondence between findings and their interval changes while jointly modeling heterogeneous trajectories across abdominal, osseous, and pulmonary regions. Although such systems may assist radiologists as decision-support tools, they cannot fully replace radiologists in clinical practice.

% ===================== Appendix E =====================
\section{Reproducibility}
\label{app:repro}

\paragraph{Artifacts.}
Upon acceptance, we will release the training code, longitudinal-pairing utilities, evaluation scripts, configuration files, and trained ALTER model weights. The release will also include the prefix-removal utilities and metric wrappers used for regional scoring. In accordance with data-use requirements and patient-privacy considerations, neither patient data nor derived image volumes will be redistributed. The main paper and this appendix jointly specify the experimental protocol required to reproduce the reported results.

\paragraph{Data access.}
RadGenome-ChestCT is derived from CT-RATE, access to which requires a PhysioNet data-use agreement. CTRG-Chest-548K is publicly available through Hugging Face. We do not redistribute any patient data. Longitudinal pairing relies exclusively on patient and study metadata available through authorized access to CT-RATE and requires no additional clinical annotations. Given these metadata, the pairing procedure strictly follows the deterministic immediate-predecessor rule described in the dataset section. This rule does not apply additional filtering based on acquisition type, contrast phase, anatomical coverage, or inter-study interval.

% ===================== BIBLIOGRAPHY =====================

\bibliography{aaai27}

@article{radgenome,
  title   = {Development of a Large-Scale Grounded Vision Language Dataset for Chest {CT} Analysis},
  author  = {Zhang, Xiaoman and Wu, Chaoyi and Zhao, Ziheng and Lei, Jiayu and Tian, Weiwei and Zhang, Ya and Xie, Weidi and Wang, Yanfeng},
  journal = {Scientific Data},
  volume  = {12},
  number  = {1},
  pages   = {1636},
  year    = {2025},
  doi     = {10.1038/s41597-025-05922-9}
}

@article{hamamci2026generalist,
  title={Generalist foundation models from a multimodal dataset for 3D computed tomography},
  author={Hamamci, Ibrahim Ethem and Er, Sezgin and Wang, Chenyu and Almas, Furkan and Simsek, Ayse Gulnihan and Esirgun, Sevval Nil and Dogan, Irem and Durugol, Omer Faruk and Hou, Benjamin and Shit, Suprosanna and others},
  journal={Nature Biomedical Engineering},
  pages={1--19},
  year={2026},
  publisher={Nature Publishing Group UK London}
}

@inproceedings{lora,
  title     = {{LoRA}: Low-Rank Adaptation of Large Language Models},
  author    = {Hu, Edward J. and Shen, Yelong and Wallis, Phillip and Allen-Zhu, Zeyuan and Li, Yuanzhi and Wang, Shean and Wang, Lu and Chen, Weizhu},
  booktitle = {International Conference on Learning Representations (ICLR)},
  year      = {2022}
}

@article{reg2rg,
  title   = {Large Language Model with Region-Guided Referring and Grounding for {CT} Report Generation},
  author  = {Chen, Zhixuan and Bie, Yequan and Jin, Haibo and Chen, Hao},
  journal = {IEEE Transactions on Medical Imaging},
  volume  = {44},
  number  = {8},
  pages   = {3139--3150},
  year    = {2025},
  doi     = {10.1109/TMI.2025.3559923}
}

@inproceedings{ct2rep,
  title={Ct2rep: Automated radiology report generation for 3d medical imaging},
  author={Hamamci, Ibrahim Ethem and Er, Sezgin and Menze, Bjoern},
  booktitle={International Conference on Medical Image Computing and Computer-Assisted Intervention},
  pages={476--486},
  year={2024},
  organization={Springer}
}

@inproceedings{pathgraphbrainct,
  title     = {Granularity Matters: Pathological Graph-Driven Cross-Modal Alignment for Brain {CT} Report Generation},
  author    = {Shi, Yanzhao and Ji, Junzhong and Zhang, Xiaodan and Qu, Liangqiong and Liu, Ying},
  booktitle = {Proceedings of the 2023 Conference on Empirical Methods in Natural Language Processing (EMNLP)},
  pages     = {6617--6630},
  year      = {2023},
  doi       = {10.18653/v1/2023.emnlp-main.408}
}

@inproceedings{hergen,
  title={Hergen: Elevating radiology report generation with longitudinal data},
  author={Wang, Fuying and Du, Shenghui and Yu, Lequan},
  booktitle={European Conference on Computer Vision},
  pages={183--200},
  year={2024},
  organization={Springer}
}

@inproceedings{comparisonprior,
  title     = {Boosting Radiology Report Generation by Infusing Comparison Prior},
  author    = {Kim, Sanghwan and Nooralahzadeh, Farhad and Rohanian, Morteza and Fujimoto, Koji and Nishio, Mizuho and Sakamoto, Ryo and Rinaldi, Fabio and Krauthammer, Michael},
  booktitle = {Proceedings of the Workshop on Biomedical Natural Language Processing and BioNLP Shared Tasks},
  pages     = {50--61},
  year      = {2023},
  doi       = {10.18653/v1/2023.bionlp-1.4}
}

@inproceedings{controllong,
    title = "Controllable Chest {X}-Ray Report Generation from Longitudinal Representations",
    author = "Dalla Serra, Francesco  and
      Wang, Chaoyang  and
      Deligianni, Fani  and
      Dalton, Jeff  and
      O{'}Neil, Alison",
    editor = "Bouamor, Houda  and
      Pino, Juan  and
      Bali, Kalika",
    booktitle = "Findings of the Association for Computational Linguistics: EMNLP 2023",
    month = dec,
    year = "2023",
    address = "Singapore",
    publisher = "Association for Computational Linguistics",
    url = "https://aclanthology.org/2023.findings-emnlp.325/",
    doi = "10.18653/v1/2023.findings-emnlp.325",
    pages = "4891--4904"
}

@article{ddatr,
  title   = {{DDaTR}: Dynamic Difference-Aware Temporal Residual Network for Longitudinal Radiology Report Generation},
  author  = {Song, Shanshan and Tang, Hui and Yang, Honglong and Li, Xiaomeng},
  journal = {IEEE Transactions on Medical Imaging},
  volume  = {44},
  number  = {12},
  pages   = {5345--5357},
  year    = {2025},
  doi     = {10.1109/TMI.2025.3591364}
}

@inproceedings{recap,
  title={RECAP: Towards precise radiology report generation via dynamic disease progression reasoning},
  author={Hou, Wenjun and Cheng, Yi and Xu, Kaishuai and Li, Wenjie and Liu, Jiang},
  booktitle={Findings of the Association for Computational Linguistics: EMNLP 2023},
  pages={2134--2147},
  year={2023}
}

@article{ctrg,
  title={Work like a doctor: Unifying scan localizer and dynamic generator for automated computed tomography report generation},
  author={Tang, Yuhao and Yang, Haichen and Zhang, Liyan and Yuan, Ye},
  journal={Expert Systems with Applications},
  volume={237},
  pages={121442},
  year={2024},
  publisher={Elsevier}
}

@inproceedings{bleu,
  title={Bleu: a method for automatic evaluation of machine translation},
  author={Papineni, Kishore and Roukos, Salim and Ward, Todd and Zhu, Wei-Jing},
  booktitle={Proceedings of the 40th annual meeting of the Association for Computational Linguistics},
  pages={311--318},
  year={2002}
}

@inproceedings{rouge,
  title={Rouge: A package for automatic evaluation of summaries},
  author={Lin, Chin-Yew},
  booktitle={Text summarization branches out},
  pages={74--81},
  year={2004}
}

@inproceedings{meteor,
  title={METEOR: An automatic metric for MT evaluation with improved correlation with human judgments},
  author={Banerjee, Satanjeev and Lavie, Alon},
  booktitle={Proceedings of the acl workshop on intrinsic and extrinsic evaluation measures for machine translation and/or summarization},
  pages={65--72},
  year={2005}
}

@inproceedings{cider,
  title={Cider: Consensus-based image description evaluation},
  author={Vedantam, Ramakrishna and Lawrence Zitnick, C and Parikh, Devi},
  booktitle={Proceedings of the IEEE conference on computer vision and pattern recognition},
  pages={4566--4575},
  year={2015}
}

@article{radgraph,
  title   = {{RadGraph}: Extracting Clinical Entities and Relations from Radiology Reports},
  author  = {Jain, Saahil and Agrawal, Ashwin and Saporta, Adriel and Truong, Steven Q.H. and Duong, Du Nguyen and Bui, Tan and Chambon, Pierre and Zhang, Yuhao and Lungren, Matthew P. and Ng, Andrew Y. and Langlotz, Curtis P. and Rajpurkar, Pranav},
  journal = {Proceedings of the Neural Information Processing Systems Track on Datasets and Benchmarks},
  year    = {2021}
}

@inproceedings{green,
  title     = {{GREEN}: Generative Radiology Report Evaluation and Error Notation},
  author    = {Ostmeier, Sophie and Xu, Justin and Chen, Zhihong and Varma, Maya and Blankemeier, Louis and Bluethgen, Christian and Michalson, Arne Edward and Moseley, Michael and Langlotz, Curtis and Chaudhari, Akshay S. and Delbrouck, Jean-Benoit},
  booktitle = {Findings of the Association for Computational Linguistics: EMNLP 2024},
  pages     = {374--390},
  year      = {2024},
  doi       = {10.18653/v1/2024.findings-emnlp.21}
}

@article{r2gengpt,
  title={R2gengpt: Radiology report generation with frozen llms},
  author={Wang, Zhanyu and Liu, Lingqiao and Wang, Lei and Zhou, Luping},
  journal={Meta-Radiology},
  volume={1},
  number={3},
  pages={100033},
  year={2023},
  publisher={Elsevier}
}

@article{medvint,
  title   = {Development of a Large-Scale Medical Visual Question-Answering Dataset},
  author  = {Zhang, Xiaoman and Wu, Chaoyi and Zhao, Ziheng and Lin, Weixiong and Zhang, Ya and Wang, Yanfeng and Xie, Weidi},
  journal = {Communications Medicine},
  volume  = {4},
  number  = {1},
  pages   = {277},
  year    = {2024},
  doi     = {10.1038/s43856-024-00709-2}
}

@misc{bai2024m3dadvancing3dmedical,
      title={M3D: Advancing 3D Medical Image Analysis with Multi-Modal Large Language Models}, 
      author={Fan Bai and Yuxin Du and Tiejun Huang and Max Q. -H. Meng and Bo Zhao},
      year={2024},
      eprint={2404.00578},
      archivePrefix={arXiv},
      primaryClass={cs.CV},
      url={https://arxiv.org/abs/2404.00578}, 
}

@inproceedings{bannur2023learning,
  title={Learning to exploit temporal structure for biomedical vision-language processing},
  author={Bannur, Shruthi and Hyland, Stephanie and Liu, Qianchu and Perez-Garcia, Fernando and Ilse, Maximilian and Castro, Daniel C and Boecking, Benedikt and Sharma, Harshita and Bouzid, Kenza and Thieme, Anja and others},
  booktitle={Proceedings of the IEEE/CVF conference on computer vision and pattern recognition},
  pages={15016--15027},
  year={2023}
}

@inproceedings{prefillrrg,
  title={Utilizing longitudinal chest x-rays and reports to pre-fill radiology reports},
  author={Zhu, Qingqing and Mathai, Tejas Sudharshan and Mukherjee, Pritam and Peng, Yifan and Summers, Ronald M and Lu, Zhiyong},
  booktitle={International Conference on Medical Image Computing and Computer-Assisted Intervention},
  pages={189--198},
  year={2023},
  organization={Springer}
}

@inproceedings{hcllm,
  title={HC-LLM: Historical-constrained large language models for radiology report generation},
  author={Liu, Tengfei and Wang, Jiapu and Hu, Yongli and Li, Mingjie and Yi, Junfei and Chang, Xiaojun and Gao, Junbin and Yin, Baocai},
  booktitle={Proceedings of the AAAI conference on artificial intelligence},
  volume={39},
  number={6},
  pages={5595--5603},
  year={2025}
}

@inproceedings{r2gen,
  title={Generating radiology reports via memory-driven transformer},
  author={Chen, Zhihong and Song, Yan and Chang, Tsung-Hui and Wan, Xiang},
  booktitle={Proceedings of the 2020 conference on empirical methods in natural language processing (EMNLP)},
  pages={1439--1449},
  year={2020}
}

@inproceedings{kerp,
  title     = {Knowledge-Driven Encode, Retrieve, Paraphrase for Medical Image Report Generation},
  author    = {Li, Christy Y. and Liang, Xiaodan and Hu, Zhiting and Xing, Eric P.},
  booktitle = {Proceedings of the AAAI Conference on Artificial Intelligence (AAAI)},
  volume    = {33},
  pages     = {6666--6673},
  year      = {2019},
  doi       = {10.1609/aaai.v33i01.33016666}
}

@inproceedings{aligntransformer,
  title     = {{AlignTransformer}: Hierarchical Alignment of Visual Regions and Disease Tags for Medical Report Generation},
  author    = {You, Di and Liu, Fenglin and Ge, Shen and Xie, Xiaoxia and Zhang, Jing and Wu, Xian},
  booktitle = {Medical Image Computing and Computer Assisted Intervention (MICCAI)},
  pages     = {72--82},
  year      = {2021},
  doi       = {10.1007/978-3-030-87199-4_7}
}

@inproceedings{rgrg,
  title={Interactive and explainable region-guided radiology report generation},
  author={Tanida, Tim and M{\"u}ller, Philip and Kaissis, Georgios and Rueckert, Daniel},
  booktitle={Proceedings of the IEEE/CVF conference on computer vision and pattern recognition},
  pages={7433--7442},
  year={2023}
}

@inproceedings{zhou2026mitigating,
  title={Mitigating entity hallucinations in 3D radiology report generation via dual-stream alignment},
  author={Zhou, Lingyu and Yu, Yue and Yi, Zhang and Xu, Xiuyuan},
  booktitle={Proceedings of the AAAI Conference on Artificial Intelligence},
  volume={40},
  number={16},
  pages={13719--13727},
  year={2026}
}

@article{ssreward,
  title   = {Longitudinal Data and a Semantic Similarity Reward for Chest {X}-Ray Report Generation},
  author  = {Nicolson, Aaron and Dowling, Jason and Anderson, Douglas and Koopman, Bevan},
  journal = {Informatics in Medicine Unlocked},
  volume  = {50},
  pages   = {101585},
  year    = {2024},
  doi     = {10.1016/j.imu.2024.101585}
}

@inproceedings{cocacxr,
  title     = {{CoCa-CXR}: Contrastive Captioners Learn Strong Temporal Structures for Chest {X}-Ray Vision-Language Understanding},
  author    = {Chen, Yixiong and Xu, Shawn and Sellergren, Andrew and Matias, Yossi and Hassidim, Avinatan and Shetty, Shravya and Golden, Daniel and Yuille, Alan L. and Yang, Lin},
  booktitle = {Medical Image Computing and Computer Assisted Intervention (MICCAI)},
  pages     = {78--88},
  year      = {2025},
  doi       = {10.1007/978-3-032-04978-0_8}
}

@inproceedings{priorrg,
  title     = {{PriorRG}: Prior-Guided Contrastive Pre-training and Coarse-to-Fine Decoding for Chest {X}-Ray Report Generation},
  author    = {Liu, Kang and Ma, Zhuoqi and Fang, Zikang and Li, Yunan and Xie, Kun and Miao, Qiguang},
  booktitle = {Proceedings of the AAAI Conference on Artificial Intelligence (AAAI)},
  volume    = {40},
  pages     = {7206--7214},
  year      = {2026},
  doi       = {10.1609/aaai.v40i9.37657}
}

@inproceedings{mare,
  title     = {{MARE}: Multimodal Analogical Reasoning for Disease Evolution-Aware Radiology Report Generation},
  author    = {Gao, Qingqing and Liu, Tengfei and Li, Xiaoyan and Zhang, Xiaodan and Sun, Zhongfan and Wang, Boyue and Yin, Baocai and Liu, Zhaohui},
  booktitle = {Proceedings of the AAAI Conference on Artificial Intelligence (AAAI)},
  volume    = {40},
  pages     = {21180--21188},
  year      = {2026},
  doi       = {10.1609/aaai.v40i25.39262}
}

@inproceedings{jing2018,
  title     = {On the Automatic Generation of Medical Imaging Reports},
  author    = {Jing, Baoyu and Xie, Pengtao and Xing, Eric},
  booktitle = {Proceedings of the 56th Annual Meeting of the Association for Computational Linguistics (ACL)},
  pages     = {2577--2586},
  year      = {2018},
  doi       = {10.18653/v1/P18-1240}
}

@inproceedings{tienet,
  title     = {{TieNet}: Text-Image Embedding Network for Common Thorax Disease Classification and Reporting in Chest {X}-Rays},
  author    = {Wang, Xiaosong and Peng, Yifan and Lu, Le and Lu, Zhiyong and Summers, Ronald M.},
  booktitle = {Proceedings of the IEEE/CVF Conference on Computer Vision and Pattern Recognition (CVPR)},
  pages     = {9049--9058},
  year      = {2018},
  doi       = {10.1109/CVPR.2018.00943}
}

@inproceedings{r2gencmn,
  title     = {Cross-Modal Memory Networks for Radiology Report Generation},
  author    = {Chen, Zhihong and Shen, Yaling and Song, Yan and Wan, Xiang},
  booktitle = {Proceedings of the 59th Annual Meeting of the Association for Computational Linguistics and the 11th International Joint Conference on Natural Language Processing (ACL-IJCNLP)},
  pages     = {5904--5914},
  year      = {2021},
  doi       = {10.18653/v1/2021.acl-long.459}
}

@misc{tian2026diffvpdifferentialvisualsemantic,
      title={DiffVP: Differential Visual Semantic Prompting for LLM-Based CT Report Generation}, 
      author={Yuhe Tian and Kun Zhang and Haoran Ma and Rui Yan and Yingtai Li and Rongsheng Wang and Shaohua Kevin Zhou},
      year={2026},
      eprint={2603.17718},
      archivePrefix={arXiv},
      primaryClass={cs.CV},
      url={https://arxiv.org/abs/2603.17718}, 
}

@article{wasserthal2023totalsegmentator,
  title={TotalSegmentator: robust segmentation of 104 anatomic structures in CT images},
  author={Wasserthal, Jakob and Breit, Hanns-Christian and Meyer, Manfred T and Pradella, Maurice and Hinck, Daniel and Sauter, Alexander W and Heye, Tobias and Boll, Daniel T and Cyriac, Joshy and Yang, Shan and others},
  journal={Radiology: Artificial Intelligence},
  volume={5},
  number={5},
  pages={e230024},
  year={2023},
  publisher={Radiological Society of North America}
}

@article{dosovitskiy2020image,
  title={An image is worth 16x16 words: Transformers for image recognition at scale},
  author={Dosovitskiy, Alexey and Beyer, Lucas and Kolesnikov, Alexander and Weissenborn, Dirk and Zhai, Xiaohua and Unterthiner, Thomas and Dehghani, Mostafa and Minderer, Matthias and Heigold, Georg and Gelly, Sylvain and others},
  journal={arXiv preprint arXiv:2010.11929},
  year={2020}
}

@article{alayrac2022flamingo,
  title={Flamingo: a visual language model for few-shot learning},
  author={Alayrac, Jean-Baptiste and Donahue, Jeff and Luc, Pauline and Miech, Antoine and Barr, Iain and Hasson, Yana and Lenc, Karel and Mensch, Arthur and Millican, Katie and Reynolds, Malcolm and others},
  journal={arXiv preprint arXiv:2204.14198},
  year={2022}
}

@article{radfm,
  title={Towards generalist foundation model for radiology by leveraging web-scale 2d\&3d medical data},
  author={Wu, Chaoyi and Zhang, Xiaoman and Zhang, Ya and Hui, Hui and Wang, Yanfeng and Xie, Weidi},
  journal={Nature Communications},
  volume={16},
  number={1},
  pages={7866},
  year={2025},
  publisher={Nature Publishing Group UK London}
}

@article{yan2022radbert,
  title={RadBERT: adapting transformer-based language models to radiology},
  author={Yan, An and McAuley, Julian and Lu, Xing and Du, Jiang and Chang, Eric Y and Gentili, Amilcare and Hsu, Chun-Nan},
  journal={Radiology: Artificial Intelligence},
  volume={4},
  number={4},
  pages={e210258},
  year={2022},
  publisher={Radiological Society of North America}
}

@article{loshchilov2017decoupled,
  title={Decoupled weight decay regularization},
  author={Loshchilov, Ilya and Hutter, Frank},
  journal={arXiv preprint arXiv:1711.05101},
  year={2017}
}

@misc{radllama7b,
  author       = {{Stanford AIMI}},
  title        = {{RadLLaMA-7b}},
  year         = {2024},
  howpublished = {\url{https://huggingface.co/StanfordAIMI/RadLLaMA-7b}},
  note         = {Hugging Face model repository. Accessed: 2026-07-29}
}

@inproceedings{cmcl,
  title     = {Competence-Based Multimodal Curriculum Learning for Medical Report Generation},
  author    = {Liu, Fenglin and Ge, Shen and Wu, Xian},
  booktitle = {Proceedings of the 59th Annual Meeting of the Association for Computational Linguistics and the 11th International Joint Conference on Natural Language Processing (ACL-IJCNLP)},
  pages     = {3001--3012},
  year      = {2021},
  doi       = {10.18653/v1/2021.acl-long.234}
}

@inproceedings{kgrrg,
  title     = {When Radiology Report Generation Meets Knowledge Graph},
  author    = {Zhang, Yixiao and Wang, Xiaosong and Xu, Ziyue and Yu, Qihang and Yuille, Alan and Xu, Daguang},
  booktitle = {Proceedings of the AAAI Conference on Artificial Intelligence (AAAI)},
  volume    = {34},
  pages     = {12910--12917},
  year      = {2020},
  doi       = {10.1609/aaai.v34i07.6989}
}

@inproceedings{ppked,
  title     = {Exploring and Distilling Posterior and Prior Knowledge for Radiology Report Generation},
  author    = {Liu, Fenglin and Wu, Xian and Ge, Shen and Fan, Wei and Zou, Yuexian},
  booktitle = {Proceedings of the IEEE/CVF Conference on Computer Vision and Pattern Recognition (CVPR)},
  pages     = {13748--13757},
  year      = {2021},
  doi       = {10.1109/CVPR46437.2021.01354}
}

\end{document}